\def\year{2022}\relax
\documentclass[letterpaper]{article} 
\usepackage{aaai22}  
\usepackage{times}  
\usepackage{helvet}  
\usepackage{courier}  
\usepackage[hyphens]{url}  
\usepackage{graphicx} 
\usepackage{natbib}  
\usepackage{caption} 
\DeclareCaptionStyle{ruled}{labelfont=normalfont,labelsep=colon,strut=off} 
\usepackage{algorithm}
\usepackage{algorithmic}

\usepackage{newfloat}
\usepackage{listings}
\floatstyle{ruled}
\newfloat{listing}{tb}{lst}{}
\floatname{listing}{Listing}
\nocopyright
\newcommand{\journal}[1]{}
\newcommand{\journalB}[1]{}
\newcommand{\journalC}[1]{}
\usepackage{hhline}
\usepackage{multirow}
\usepackage{MnSymbol}
\usepackage{subcaption}
\usepackage{mathtools}

\usepackage{color}
\definecolor{Blue}{rgb}{0,0,1}

\definecolor{Orange}{rgb}{1,0.5,0}

\definecolor{Green}{rgb}{0,1,0}

\newlength{\tempheight}
\newlength{\tempwidth}
\newcommand{\rowname}[1]
{\rotatebox{90}{\makebox[\tempheight][c]{\small \textbf{#1}}}}
\newcommand{\columnname}[1]
{\makebox[\tempwidth][c]{\small \textbf{#1}}}

\title{Fairness-Driven Private Collaborative Machine Learning}
\author {
    Dana Pessach\textsuperscript{\rm 1},
    Tamir Tassa\textsuperscript{\rm 2},
    Erez Shmueli\textsuperscript{\rm 1,*}
}
\affiliations {
    \textsuperscript{\rm 1}Department of Industrial Engineering, Tel-Aviv University, Israel\\
    \textsuperscript{\rm 2}Department of Mathematics and Computer Science, The Open University, Israel\\
    \textsuperscript{\rm *}Corresponding author: shmueli@tau.ac.il
}

\begin{document}

\maketitle

\begin{abstract}
The performance of machine learning algorithms can be considerably improved when trained over larger datasets.
In many domains, such as medicine and finance, larger datasets can be obtained if several parties, each having access to limited amounts of data, collaborate and share their data.
However, such data sharing introduces significant privacy challenges.
While multiple recent studies have investigated methods for private collaborative machine learning, the fairness of such collaborative algorithms was overlooked.
In this work we suggest a feasible privacy-preserving pre-process mechanism for enhancing fairness of collaborative machine learning algorithms.
Our experimentation with the proposed method shows that it is able to enhance fairness considerably with only a minor compromise in accuracy.
\journal{
An extensive evaluation of the proposed method shows that it is able to enhance fairness considerably with only a minor compromise in accuracy.
}
\end{abstract}

\section{Introduction}
\journalB{
Machine learning (ML) is one of the most prominent technological developments in recent years, rapidly becoming a central part of our life.
Applications of ML are all around us, ranging from traffic prediction and virtual personal assistants to automated radiology and autonomous vehicles.
}

The performance of machine learning (ML) models inherently depends on the availability of a large quantity of useful training data.
In neural network learning, for example, recent studies have shown that accuracy can be improved considerably by having access to large datasets \cite{krizhevsky2012imagenet}. 
In many applications, however, data is scattered and held by multiple different parties that may be reluctant to share their information for multiple reasons such as commercial competition, privacy concerns, and in some domains even legal constraints.
For example, the Health Insurance Portability and Accountability Act (HIPAA) in the United States
places strict constraints on the ability of health care providers to share patient data \cite{HIPAA:1996:Online}.

As a result, a variety of methods have been recently proposed to allow multiple parties to collaboratively train ML models while preventing the disclosure of private information.
Hereinafter, we refer to this setting as \textit{Private Collaborative Machine Learning} (PCML).
While these methods address a very similar problem, they are often associated with different (though overlapping) research domains including \textit{privacy-preserving data mining} \cite{lindell2000privacy,agrawal2000privacy}, \textit{privacy-preserving machine learning} \cite{jayaraman2018distributed}, \textit{collaborative machine learning} \cite{chase2017private,hu2019fdml}, and \textit{federated machine learning} \cite{yang2019federated}.

These methods take two main approaches for preserving privacy.
The first approach is based on \textit{perturbation} \cite{agrawal2000privacy}; it incorporates noise to the training data in order to obscure sensitive information. 
The main limitation of this approach is that incorporating noise to the data may yield an inferior model.
The second approach is based on \textit{Secure Multi-party Computation} (SMC) \cite{lindell2000privacy}.
SMC achieves privacy protection by applying cryptographic techniques that enable several parties to perform a joint computation on private data that is distributed among them, where only the outcome of the
computation is revealed to the parties, but no other information is exposed. 
In contrast to perturbation, SMC does not change the data, and therefore the output issued by such algorithms is identical to the output of their non-privacy-preserving counterparts.
However, since in many cases, a generic and perfectly secure SMC solution is infeasible \cite{domingo2020privacy}, lower security requirements are typically accepted for the sake of higher efficiency \cite{du2004privacy}.

Despite the growing number of studies proposing algorithms for PCML, the fairness of such algorithms was overlooked.
Since many automated decisions (including which individuals will receive jobs, loans, medication, bail or parole) can significantly impact peoples' lives, there is great importance in assessing and improving the fairness of the decisions made by such algorithms.
Indeed, in recent years, the concern for algorithmic fairness has made headlines.
The evidence of algorithmic bias 
(see e.g. \citeauthor{Angwin:2016:Online} \citeyear{Angwin:2016:Online})
and the
resulting concerns for algorithmic fairness have led to growing interest in the literature on defining, evaluating and improving fairness in ML algorithms \cite[see a recent review in][]{pessach2020algorithmic}.
All of these studies, however, focused on centralized settings.

In this paper, we consider a learning setting similar to the PCML setting described above, where data is scattered among several parties, who wish to engage in a joint ML procedure, without disclosing their private information.
Our setting, however, also adds a fairness requirement, mandating that the learnt model satisfies a certain level of fairness.

To address the new fairness requirement, we suggest a privacy-preserving pre-process mechanism for enhancing fairness of collaborative ML algorithms.
Similarly to the pre-process fairness mechanism suggested in \cite{feldman2015certifying}, our method improves fairness through decreasing distances between the distributions of attributes in the privileged and unprivileged groups.
As a pre-process mechanism, our approach is not tailored to a specific algorithm, and therefore can be used with any PCML algorithm.
In contrast to \cite{feldman2015certifying}, our method was designed to allow privacy-preserving enhancements, which are obtained through SMC techniques.

Experimentation that we conducted over a real-world dataset shows that the proposed method is able to improve fairness considerably, with almost no compromise in accuracy.
Furthermore, we show that the runtime of the proposed method is feasible, especially considering that it is executed once as a pre-process procedure.

\journal{
An extensive evaluation that we conducted over real-world datasets shows that the proposed method is able to improve fairness considerably, with almost no compromise in accuracy.
Furthermore, we show that the runtime of the proposed method is feasible, especially considering that it is executed once as a pre-process procedure.
}


\section{Related Work}
\label{S:related_work}

We organize the survey of relevant related work as follows.
In Section \ref{S:algof} we survey related
studies that deal with algorithmic fairness in centralized settings. In Section \ref{S:22} we review literature on private collaborative machine learning. We conclude, in Section \ref{S:23}, with a discussion of recent research efforts on private and fair machine learning. We note that all of the latter studies focused on either a centralized setting or a distributed setting which is non-collaborative, while we consider here a distributed collaborative setting.

\subsection{Algorithmic Fairness in Centralized Settings}
\label{S:algof}

\journalC{\subsection*{\textit{Fairness Definitions and Measures}}}

\journal{
The law makes a distinction between two types of discrimination:
i) \textit{Disparate treatment} \citep{zimmer1995emerging,barocas2016big}: intentionally treating an individual differently based on his/her membership in a protected class/unprivileged group (\textit{direct discrimination}); and
ii) \textit{Disparate impact}  \citep{rutherglen1987disparate,barocas2016big}: negatively affecting members of a protected class/privileged group, more than others, even if by a seemingly neutral policy (\textit{indirect discrimination}).

Put in our context, it is important to note that algorithms trained with data that do not include sensitive attributes (i.e., attributes that explicitly identify the privileged and unprivileged groups) are unlikely to produce \textit{disparate treatment}, but may still induce unintentional discrimination in the form of \textit{disparate impact} \citep{kleinberg2017inherent}.
}

In the algorithmic fairness literature, multiple measures were suggested.
 \cite[The reader is referred to][for a comprehensive review of fairness definitions and measures]{pessach2020algorithmic,verma2018fairness}.
The most prominent measures include \textit{demographic parity} \cite{calders2010three,dwork2012fairness} and \textit{equalized odds} \cite{hardt2016equality}.
Demographic parity ensures that the proportion of the positive predictions is similar across groups.
\journalC{For example, if a positive prediction represents acceptance for a job, then the demographic parity condition requires the proportion of accepted applicants to be similar across groups.}
One disadvantage of this measure is that a fully accurate classifier may be considered unfair, when the base rates (i.e., the proportion of actual positive outcomes) of the various groups are considerably different.
Moreover, in order to satisfy {demographic parity}, two similar individuals may be treated differently since they belong to two different groups, and in some cases, such treatment is prohibited by law.
\journalC{}
In this paper we focus on a variation of the {equalized odds} measure.
This measure was devised by \citeauthor{hardt2016equality} \shortcite{hardt2016equality} to overcome the disadvantages of measures such as {demographic parity}.
It was designed to assess the difference between the two groups by measuring the difference between the false positive rates (FPR) in the two groups, as well as the difference between the false negative rates (FNR) in the two groups.
\journalC{}
In contrast to the {demographic parity} measure, a fully accurate classifier will necessarily satisfy the two {equalized odds} constraints.
Nevertheless, since {equalized odds} relies on the actual ground truth, it assumes that the base rates of the two groups are representative and were not obtained in a biased manner.
\journalC{}
It is important to note that there is an inherent trade-off between accuracy and fairness: as we pursue a higher degree of fairness, we may compromise accuracy 
\cite[see, for example,][]{kleinberg2017inherent}. 

\journalC{\subsection*{Mechanisms for Enhancing Fairness}}

Fairness-enhancing mechanisms are broadly categorized into three types:
pre-process, in-process and post-process.
Pre-process mechanisms involve changing the training data before feeding it into the ML algorithm
\cite[e.g.][]{louizos2017variational,feldman2015certifying,calmon2017optimized,zemel2013learning}.
\journalB{
\cite{kamiran2012data,luong2011k,louizos2017variational,feldman2015certifying,calmon2017optimized,zemel2013learning}.
} 
In-process mechanisms involve modifying ML algorithms to account for fairness during training time 
\cite[such as][]{kamishima2012fairness,zafar2017fairness,louizos2017variational,agarwal2018reductions}.
\journalB{
\citep{kamishima2012fairness,woodworth2017learning,bechavod2017learning,bechavod2017penalizing,zafar2017fairness,zafar2017fairnessAIAS,kamiran2010discrimination,zemel2013learning,louizos2017variational,quadrianto2017recycling,goh2016satisfying,calders2010three,agarwal2018reductions}.
} 
Post-process mechanisms perform post-processing to the output scores of the model to make decisions fairer \cite{corbett2017algorithmic, dwork2018decoupled,hardt2016equality,menon2018cost}.
\journalB{
Similarly, \citeauthor{corbett2017algorithmic} \shortcite{corbett2017algorithmic} suggested to select a threshold for each group separately, in a manner that maximizes accuracy and minimizes demographic parity.
} 
The reader is referred to \cite{pessach2020algorithmic} for further information regarding different fairness-enhancing mechanisms.
\journalC{
}
\journalB{
The different mechanism types are associated with the following advantages and disadvantages \citep{pessach2020algorithmic}.
Pre-process mechanisms can be advantageous since they can be used with any ML algorithm.
However, since they are not tailored to a specific algorithm, there is no guarantee on the level of accuracy that such a mechanism may yield at the end of the process.
Similar to pre-process mechanisms, post-process mechanisms may be used with any ML algorithm.
However, due to the relatively late stage in the learning process in which they are applied, post-process mechanisms typically obtain inferior results \citep{woodworth2017learning}.
Another danger of post-process mechanisms is that they may treat entirely differently two individuals who are very similar across all attributes except for the group to which they belong.
In-process mechanisms are beneficial since they can explicitly impose the required trade-off between accuracy and fairness in the objective function.
However, such mechanisms are tightly coupled with the ML algorithm.
} 
\journalC{}
Note that the above mentioned papers focused on a centralized setting, in which all information is held by one party, and therefore did not have to deal with privacy considerations.

\subsection{Private Collaborative Machine Learning}
\label{S:22}

\journalB{
\subsection*{The Setting}
} 

We consider a setting in which several parties wish to collaboratively train ML models on data that is distributed among them, while preventing the disclosure of private information.
We refer to this setting as \textit{Private Collaborative Machine Learning} (PCML).
While many methods were proposed in the literature to address this setting (or a very similar one), they are often associated with different (though overlapping) research domains including \textit{privacy-preserving data mining} \cite{lindell2000privacy,agrawal2000privacy}, \textit{privacy-preserving machine learning} \cite{jayaraman2018distributed}, \textit{collaborative machine learning} \cite{chase2017private,hu2019fdml}, and \textit{federated machine learning} \cite{yang2019federated}.

The manner in which data is distributed among the collaborating parties is typically categorized as vertical, horizontal or mixed.
In a vertical distribution, each party holds all records with a different subset of attributes, whereas in a horizontal distribution each party holds a subset of the records with all attributes.
The mixed scenario refers to an arbitrary partition of the data among the collaborating parties.

When analyzing the privacy preservation of the protocol,
it is common to distinguish between two types of adversaries: semi-honest and malicious.
A semi-honest adversary is a party that follows the protocol properly, but tries to infer sensitive information of other parties from intermediate messages that are received during the protocol.
A malicious adversary, on the other hand, may deviate from the prescribed protocol in attempt to learn sensitive information on others.
In addition, such a malicious adversary can corrupt more than just one party, in order to increase the data at its disposal.
The semi-honest model is often more realistic and very common in the literature.

The two main approaches for achieving privacy in PCML
are perturbation and SMC (secure multi-party computation).
Perturbation-based PCML methods were described in e.g. 
\citeauthor{mangasarian2008privacy} \shortcite{mangasarian2008privacy},
\citeauthor{jeckmans2012privacy} \shortcite{jeckmans2012privacy},
\citeauthor{song2013stochastic} \shortcite{song2013stochastic}, and
\citeauthor{abadi2016deep} \shortcite{abadi2016deep}.
SMC-based PCML methods were described in e.g.
\citeauthor{jha2005privacy} \shortcite{jha2005privacy},
\citeauthor{jayaraman2018distributed} \shortcite{jayaraman2018distributed}, and
\citeauthor{TassaGY21} \shortcite{TassaGY21}.

\subsection{Privacy and Fairness}\label{S:23}

While many recent studies have investigated algorithms for PCML, the fairness of such algorithms was overlooked.
There were, however, several recent studies that incorporated both privacy and fairness considerations to ML algorithms in non-collaborative settings.
Most of these studies have investigated the case of a centralized setting, in which all of the dataset is held by a single party \cite{jagielski2019differentially,xu2019achieving,mozannar2020fair,bagdasaryan2019differential,huang2018generative,cummings2019compatibility}.
The goal of those studies was to train an ML model over the centralized dataset, making sure that the released model and its future outputs are fair, as well as private, in the sense that one cannot infer from them (meaningful) information about individual data records of the dataset. 

Few other studies have investigated 
a distributed setting
which is non-collaborative in the following sense.
Their distributed setting includes a ``main'' party that wishes to train a fair ML model over the data it holds, and a third party to which the sensitive attributes are outsourced.
The motivation behind this setting is that while the sensitive attributes should not be exposed to the main party, they should still be used in the training process of the ML model (in a privacy-preserving manner) to ensure that the resulting model is fair.
To obtain this goal, these studies used either random projections \cite{hu2019distributed,fantin2020distributed} or SMC techniques \cite{kilbertus2018blind}.

\journalC{It is worth noting that the term ``fair'' is also used in the cryptographic literature in a totally different sense.
An SMC protocol is considered fair, in that context, if it ensures that either all parties receive their designated outputs, or none of them does \cite{domingo2020privacy}. As explained earlier, in Section \ref{S:algof}, our notion of fairness is different from the above mentioned SMC fairness.}

\section{The Proposed Method}
\label{S:methods}

We propose a private pre-process fairness-enhancing mechanism based on SMC for the PCML setting.
Our solution assumes that data is horizontally distributed between parties and that adversaries are
semi-honest and non-colluding.

\subsection{Terminology and Notations}\label{S31}

Let $D$ be a dataset that is distributed horizontally among $L$ parties, $P_\ell$, $\ell \in [L]$. (Hereinafter, if $N$ is any integer then $[N]:=\{1,\ldots,N\}$.) They wish to engage in a  {\it fair} collaborative
ML classification process, without disclosing the private information that
each of them holds. To that end, we develop herein a pre-process mechanism for enhancing fairness in distributed settings,
and then we devise a privacy-preserving implementation of that mechanism, using SMC.

We proceed to introduce the basic notations that will be used throughout the paper. Other notations will be introduced later on. \journal{
A summary of all notations is provided in Appendix \ref{app:notations}.
}
%
Let $W$ represent a given population consisting of $n$ individuals, and let $A$ be a set of {\it attributes} or {\it features} that relate to each of those individuals. Then we consider a dataset $D$ in which there is a row for every individual in $W$ and a column for every attribute in $A$.
We distinguish between three types of attributes (columns):
\begin{itemize}
    \item $S$ represents a sensitive attribute (e.g., race or gender). In this work we focus on a binary attribute $S$ that attains one of two possible values, $S \in \{U,V\}$, where $U$ means {\it unprivileged} and $V$ means {\it privileged}. (By abuse of notation, we are using $S$ to denote the attribute, as well as its values in different rows of the dataset.)
    \item $X$ stands for the collection of all non-sensitive attributes. To simplify our discussion, we assume that $X$ consists of one attribute. When $X$ consists of several non-sensitive attributes, we will apply the same pre-process mechanism on each such attribute, independently. We shall also assume hereinafter that $X$ is a numerical attribute.
    \item $Y$ is the binary class attribute that needs to be predicted (e.g. "hire/no hire").
\end{itemize}

\noindent
The set of rows, or $W$, is split in two different manners.
The first split is as induced by the sensitive attribute $S$. Namely, $W = W^U \bigcupdot W^V$ (hereinafter $\bigcupdot$ denotes a {\it disjoint} union), where $W^U$ is the subset of all individuals in $W$ that are unprivileged (i.e., $S=U$ for them) and $W^V = W \setminus W^U$ is the complementary set of privileged individuals. 
For each $S \in \{U,V\}$ we let $n^S:=|W^S|$.

The other manner in which $W$ is split is according to the distribution of the records of $D$ among the $L$ parties. Namely, $W = \bigcupdot_{\ell \in [L]} W_\ell$, where $W_\ell$ is the subset of individuals whose information is held by the party
$P_\ell$, $\ell\in [L]$.
For each $\ell \in [L]$ we let $n_\ell:=|W_\ell|$.

Finally, we let $n_\ell^S$ denote the size of $W_\ell^S:=W^S \bigcap W_\ell$, namely, the number of individuals in $W^S$ whose records are held by $P_\ell$, $S \in \{U,V\}$, $\ell \in [L]$.
Hence, in summary,
\begin{equation} 
n = \sum_{S \in \{U,V\} } n^S = \sum_{\ell \in [L]} n_\ell = \sum_{ S \in \{U,V\}} \sum_{\ell \in [L]} n^S_\ell \,.
\label{nequ}
\end{equation}

We shall adopt these superscript and subscript conventions hereinafter. Namely, a superscript $S$ will denote a restriction to the subset $W^S$, $S \in \{U,V\}$; a subscript $\ell$ will denote a restriction to the subset $W_\ell$, $\ell \in [L]$; a combination of the two will denote a restriction to $W_\ell^S$; and no superscript $S$ nor subscript $\ell$ relates to the entire population $W$.  

In addition, we let $D(X)$ denote the collection of all values appearing in the $X$-column of $D$. Similarly, $D^S(X)$, $D_\ell(X)$ and $D_\ell^S(X)$ denote the collection of all values in the $X$-column of $D$, restricted to the rows in $W^S$, $W_\ell$ or $W_\ell^S$, respectively.
We assume that $D^S_\ell(X)$, for any $S$ and $\ell$, are multisets; namely, they may contain repeated values.

The rest of Section \ref{S:methods} is organized as follows. In Section \ref{S:preprocess_mechanism} we introduce our pre-process fairness-enhancing mechanism. We do that for a centralized setting, in which all information is held by one party (i.e., $L=1$); in such a setting privacy is irrelevant.
Then, in Section \ref{S:33} we show how to implement that fairness mechanism in the (horizontally) distributed setting, $L>1$, in a manner that offers privacy to the interacting parties $P_\ell$, $\ell \in [L]$.
\journal{
We conclude with analyses of privacy and computational costs of our algorithm in Sections \ref{S:34} and \ref{computational_costs}.
}

\subsection{A Pre-Process Fairness-Enhancing Mechanism}
\label{S:preprocess_mechanism}
Inspired by \citeauthor{feldman2015certifying} \shortcite{feldman2015certifying}, we devise a methodology that improves fairness through decreasing distances between the distributions of attributes of the privileged and unprivileged groups, in a pre-process stage.
The goal in performing such a repair is to reduce the dependency of the ML model on the group to which an individual belongs, even when that dependency is not a direct one but rather an indirect one through proxy variables. That is, we 
reduce the ability to differentiate between groups using the presumably legitimate non-sensitive variables.
In contrast to \citeauthor{feldman2015certifying} \shortcite{feldman2015certifying}, our method is designed specifically to be suitable for privacy-preserving enhancements (see Section \ref{S:33}). 

To illustrate the intuition behind the proposed method, consider the example depicted in Figure \ref{fig:example}.
The figure illustrates a case in which SAT scores of individuals are used to predict their success (or failure) in a given job.
The plot on the left shows the distribution of SAT scores, within the two sub-populations in the dataset --- the privileged group and the unprivileged group.
The plot on the right shows the job success rate 
as a function of the SAT score, within each of those two sub-populations.

\journalC{
\begin{figure*}[t]
\centering
\begin{subfigure}{1\linewidth}
  \centering
  \includegraphics[width=0.7\linewidth]{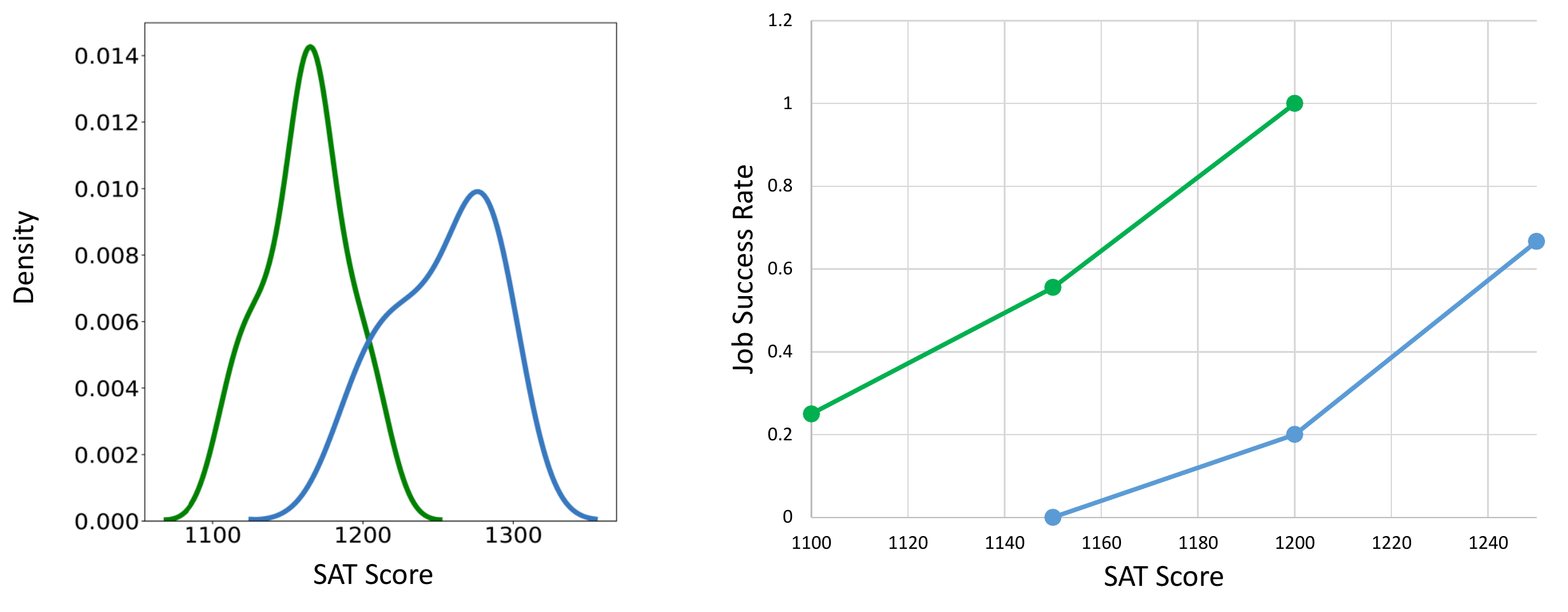}  
  \label{fig:example_sub1}
\end{subfigure}
\begin{subfigure}{.3\linewidth}
  \includegraphics[width=0.7\linewidth]{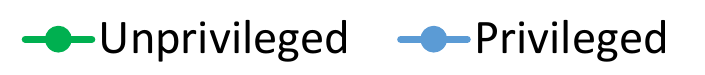}  
  \label{fig:example_sub2}
\end{subfigure}
\caption{The distribution of SAT scores within each of the two sub-populations (left), and the success rate as a function of the SAT score, within each of those sub-populations (right).}
\label{fig:example}
\end{figure*}
}

\begin{figure}[t]
\centering
\includegraphics[width=0.9\columnwidth]{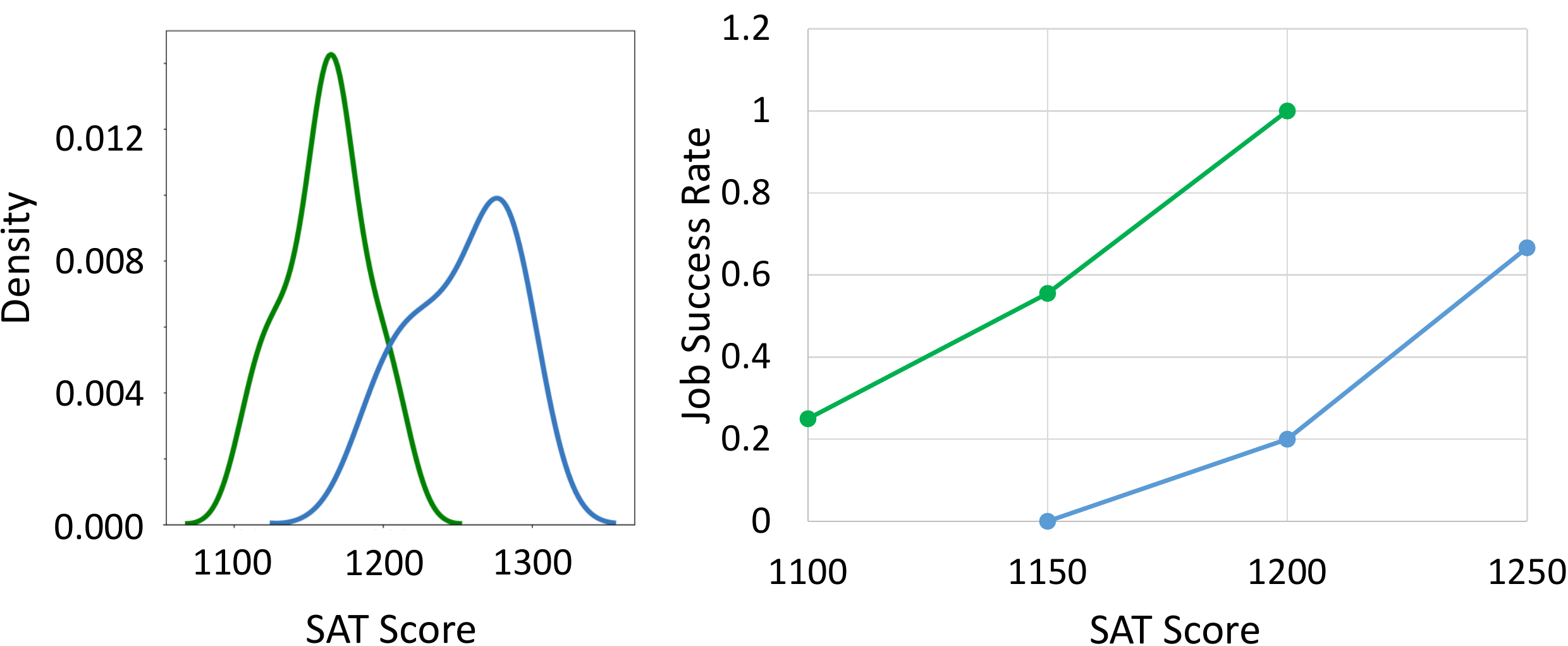}
\includegraphics[width=0.45\columnwidth]{Figures/leg_priv_unpriv.pdf}  
\caption{The distribution of SAT scores within each of the two sub-populations (left), and the success rate as a function of the SAT score, within each sub-population (right).}
\label{fig:example}
\end{figure}

In this example, SAT scores may be used to predict success in a given job, since the higher the SAT score is, the higher is the probability for success. However, relying solely on the SAT scores, while ignoring the group to which the candidates belong, may create an undesired bias.
More specifically, as can be seen from the figure, unprivileged candidates with SAT scores of approximately 1100 perform just as well as privileged candidates with SAT scores of 1200 (could be because they may have encountered harder challenges along their way for achieving their scores).
Therefore, if SAT scores were used for hiring, say by just placing a threshold, unprivileged candidates with high potential would be excluded, whereas lower potential privileged candidates would be hired instead.
The goal of the pre-process mechanism that we suggest herein is to rectify this bias.

For each group $S \in \{U,V\}$, we first partition the multiset of values in $D^S(X)$ into (nearly) equal-sized bins of sequential values. Namely, if we let $B$ denote the number of bins, then we partition $D^S(X)$ into
$D^S(X) = \bigcupdot_{i=1}^B b_{(i)}^S$,
where the following two conditions hold:\footnote{We use subscripts in parentheses for any indexing purpose that does not relate to the distribution between the $L$ parties.}
(a) for all $i \in [B-1]$, $\beta \in b^S_{(i)}$ and
$\beta' \in b^S_{(i+1)}$, we have $\beta \leq \beta'$;
and (b)
the bins are of (nearly) equal sizes, in the sense that the sizes of the bins (viewed as multisets) are either $\lfloor n^S/B \rfloor$ or $\lceil n^S/B \rceil$.
These two conditions simply state that the resulting bins are the $B$-quantiles of $D^S(X)$. Below we provide the precise manner in which those bins/quantiles are computed.

After completing the binning process, we compute the minimal values in each of the
$B$ bins, for each group $S \in \{U,V\}$,
\begin{equation}
    m^S_{(i)} := \min \{ b^S_{(i)} \} \,, \quad S \in \{U,V\}\,, ~~ i \in [B]\,, \label{midef} 
\end{equation}
and also the maximal value in the last bin,
\begin{equation}
    m^S_{(B+1)} := \max \{ b^S_{(B)} \} \,, \quad S \in \{U,V\}\,. \label{mB1def} 
\end{equation}
Those values enable us to perform a repair of $D^V(X)$, as follows: for every bin $ b^V_{(i)}$ in $D^V(X)$, $i \in [B]$,
we shift all values in it ``towards" the values in the corresponding bin in $D^U(X)$, according to the following repair rule:


\begin{equation}
\label{eq:repproposedII}
\begin{aligned}
& \bar{x}=\left(1-\lambda\right)\cdot x + \\
& +\lambda\cdot\left(m_{(i)}^U+\left(\frac{x-m_{(i)}^V}{m_{(i+1)}^V-m_{(i)}^V}\right)\cdot \left( m_{(i+1)}^U-m_{(i)}^U \right) \right), \\
& ~~~ i \in [B], x\in b_{(i)}^V\,.
\end{aligned}
\end{equation}

\noindent
Here, $x$ is an original value extracted from the bin $b_{(i)}^V$, while $\bar{x}$ is its repaired value. 
This repair procedure represents a linear mapping that performs min-max scaling of the values in each bin of the privileged group to the range of values in the corresponding bin in the unprivileged group. The computation brings the distributions of both groups closer.

The repair tuning parameter $\lambda \in [0,1]$ controls the strength of the repair. If $\lambda=0$ we perform no repair, since then $\bar{x}=x$. If $\lambda=1$ then we get a full repair, since then all values in $b_{(i)}^V$ are replaced with values in the range of the corresponding bin $b_{(i)}^U$, while keeping a similar distribution as the original ones in $b_{(i)}^V$.
Note that if all values in $b_{(i)}^V$ are equal, then both the numerator and denominator in the fraction on the right hand side of Eq. (\ref{eq:repproposedII}) are zero. In such a case we interpret that fraction as $1/2$. Such a setting implies that all the (equal) values in $b_{(i)}^V$
will be mapped to the same value in the middle of the range of the corresponding bin $b_{(i)}^U$, namely 
$\left( m_{(i+1)}^U+m_{(i)}^U \right)/2$.

At the completion of the repair pre-process procedure, an ML model is trained with the repaired dataset.


\medskip
To conclude this section, we provide the formal details of the above-described
binning scheme. Let us first order all values in $D^S(X)$, the size of which is $n^S$, in a non-decreasing manner, as follows:  
\begin{equation}
    D^S(X)= (x_1,\ldots,x_{n^S})\,, \ \mbox{where}~~
    x_1 \leq x_2 \leq \cdots \leq x_{n^S}\,. \label{DSXorder}
\end{equation}
Next, we will define a set of indices in the ordered multiset $D^S(X)$ in the following manner:
\begin{equation}
    K^S :=\{k_{(0)}:=0< k_{(1)} < \cdots < k_{(B-1)} < k_{(B)}:=n^S\} \,.\label{KS}
\end{equation}
To that end, let $q^S$ and $r^S$ be the quotient and remainder, respectively, when dividing $n^S$ by $B$; i.e.,
$n^S=q^S\cdot B+r^S$, where $r^S \in [0,B)$.
Then the sequence of indices $k_{(i)}$, $i \in [B]$, is defined by the following equation,
\begin{equation}
\begin{aligned}
& k_{(0)} = 0 \,; \quad 
    k_{(i)}=k_{(i-1)}+\Delta\,, 
    ~~ 
    \mbox{where} \\
    & ~~\Delta=\left\{\begin{matrix}
q^S+1 \quad \mbox{if} \quad i\leq r^S \\ 
q^S \quad \mbox{if} \quad i > r^S 
\end{matrix}\right. \,, 
 ~~~i \in [B]\,.\label{kidef}
\end{aligned}
    \end{equation}
It is easy to verify that $k_{(B)}=n^S$. Finally, the bins in $D^S(X)$, Eq. (\ref{DSXorder}), are defined by $K^S$ as follows:
\begin{equation}
    b^S_{(i)}:=\{ x_{j}: k_{(i-1)}+1 \leq j \leq k_{(i)}\} \,, \quad i \in [B]\,. \label{bindef}
\end{equation}
We see, in view of Eq. (\ref{bindef}), that
the size of the first $r^S$ bins is $q^S+1$, while all other bins are of size $q^S$.
Moreover, $m^S_{(i)} = \min \{ b^S_{(i)} \}$ (Eq. (\ref{midef})) equals the $(k_{(i-1)}+1)$-th ranked element
in $D^S(X)$, for all $i \in [B]$ and $S\in \{U,V\}$,
while
$  m^S_{(B+1)} := \max \{ b^S_{(B)} \}$ (Eq. (\ref{mB1def})),
equals the maximal element in $D^S(X)$, $S\in \{U,V\}$,

\begin{figure*}[t]
\settoheight{\tempheight}{\includegraphics[width=0.32\linewidth]{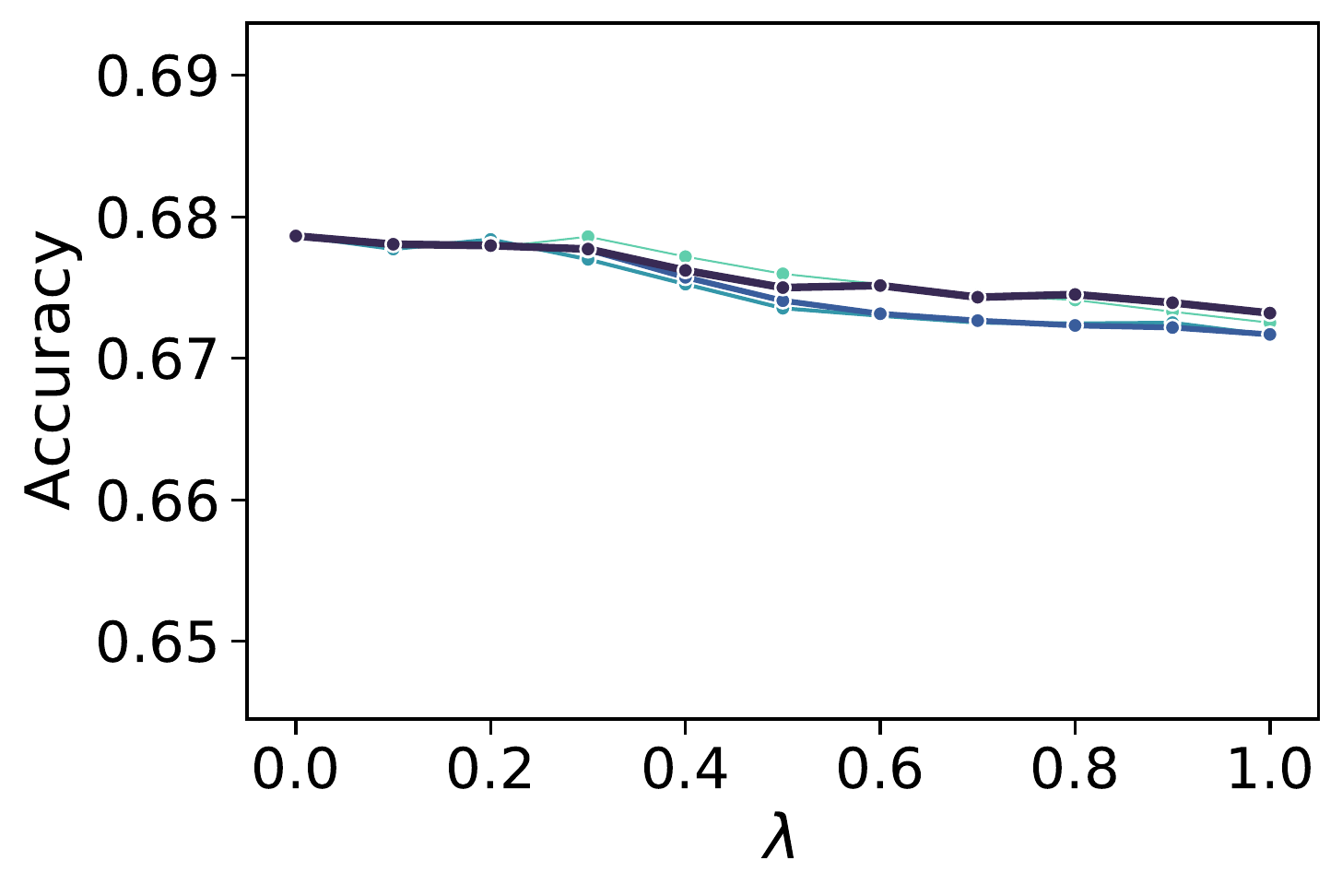}}
\centering
\columnname{Distance}\hfil\hfil
\columnname{Unfairness}\hfil\hfil
\columnname{Accuracy}\hfil\hfil
\\
\vspace{1mm}
{\includegraphics[width=0.305\linewidth]{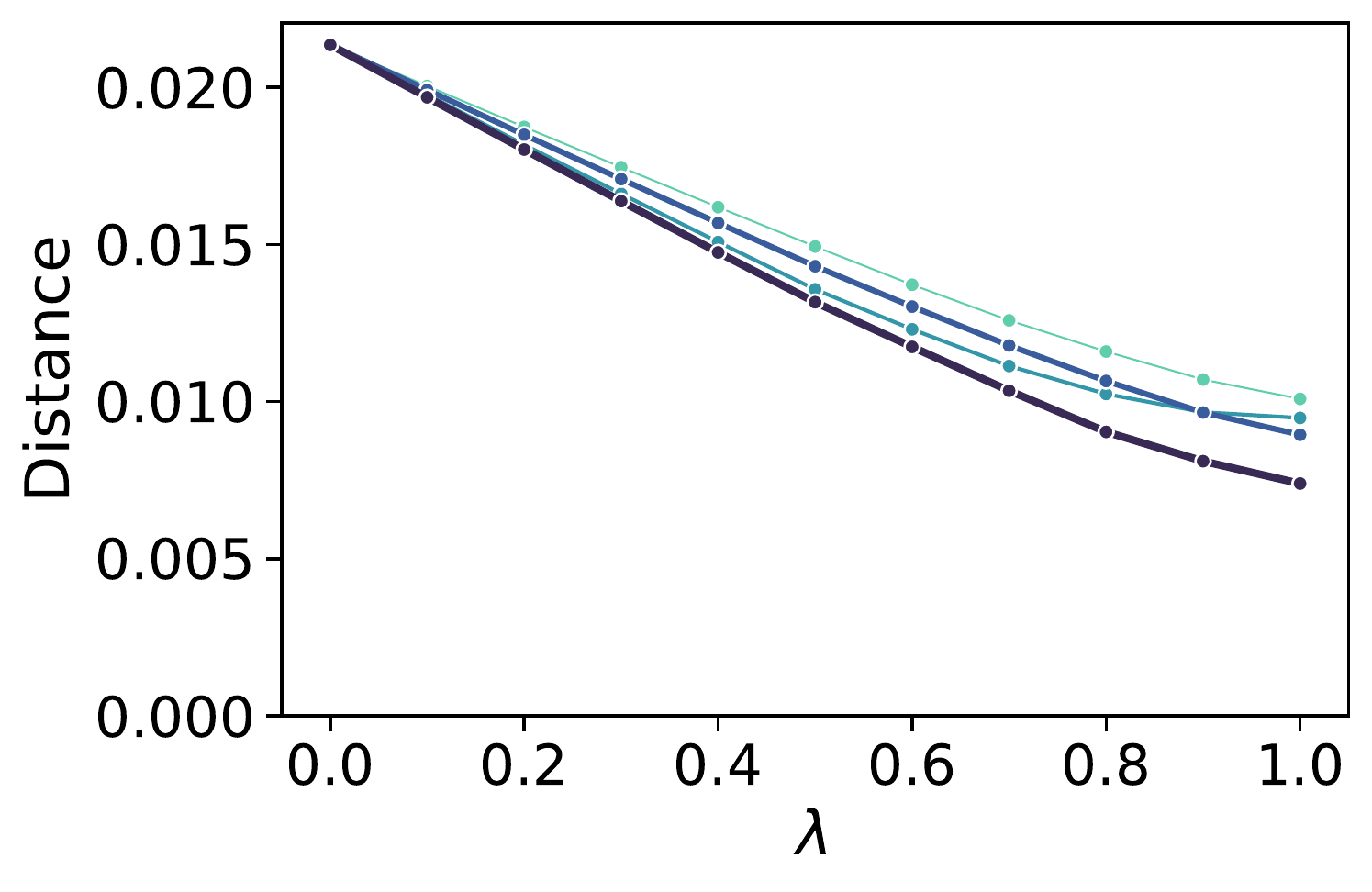}}
\hfil
{\includegraphics[width=0.29\linewidth]{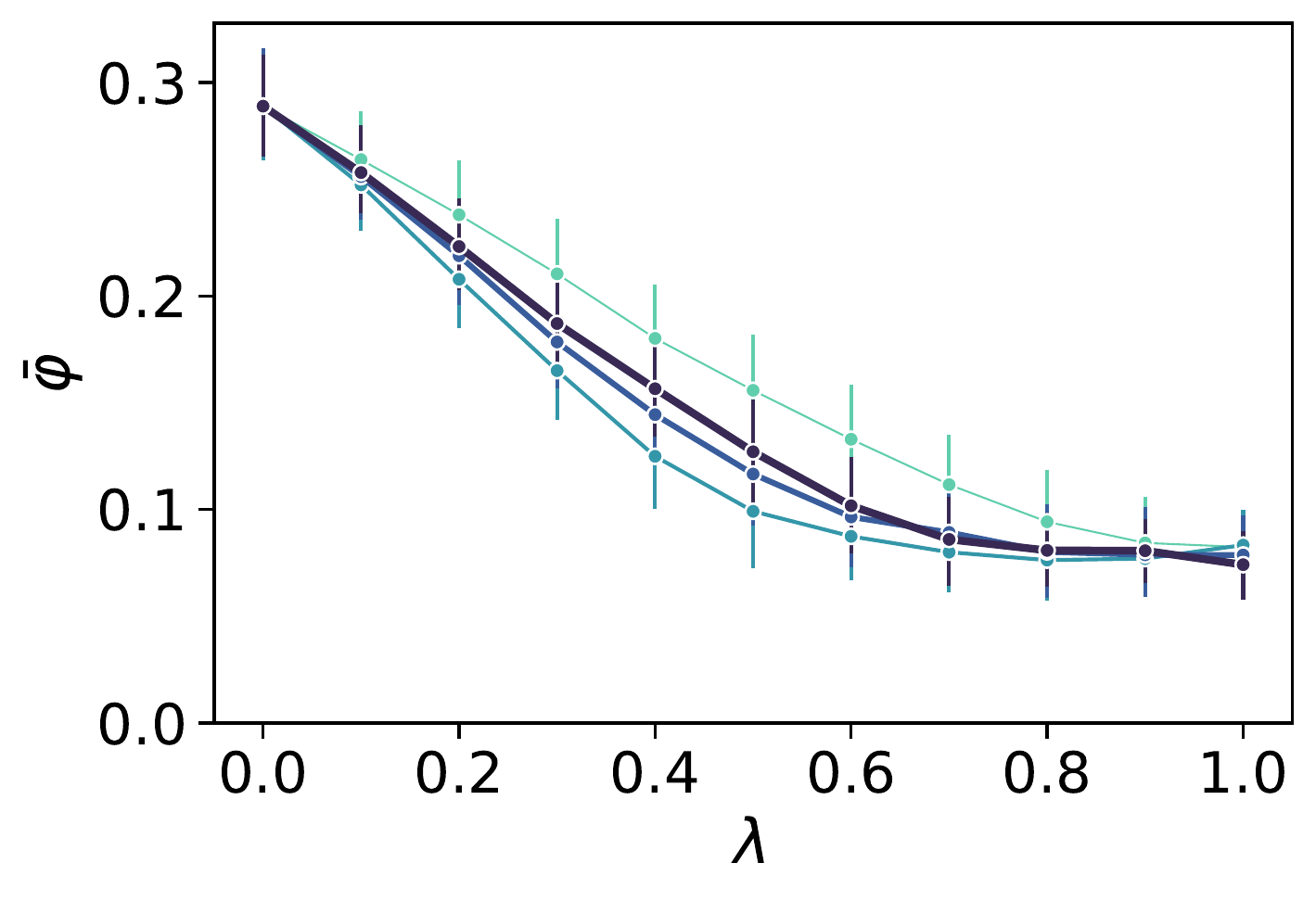}}
\hfil
{\includegraphics[width=0.3\linewidth]{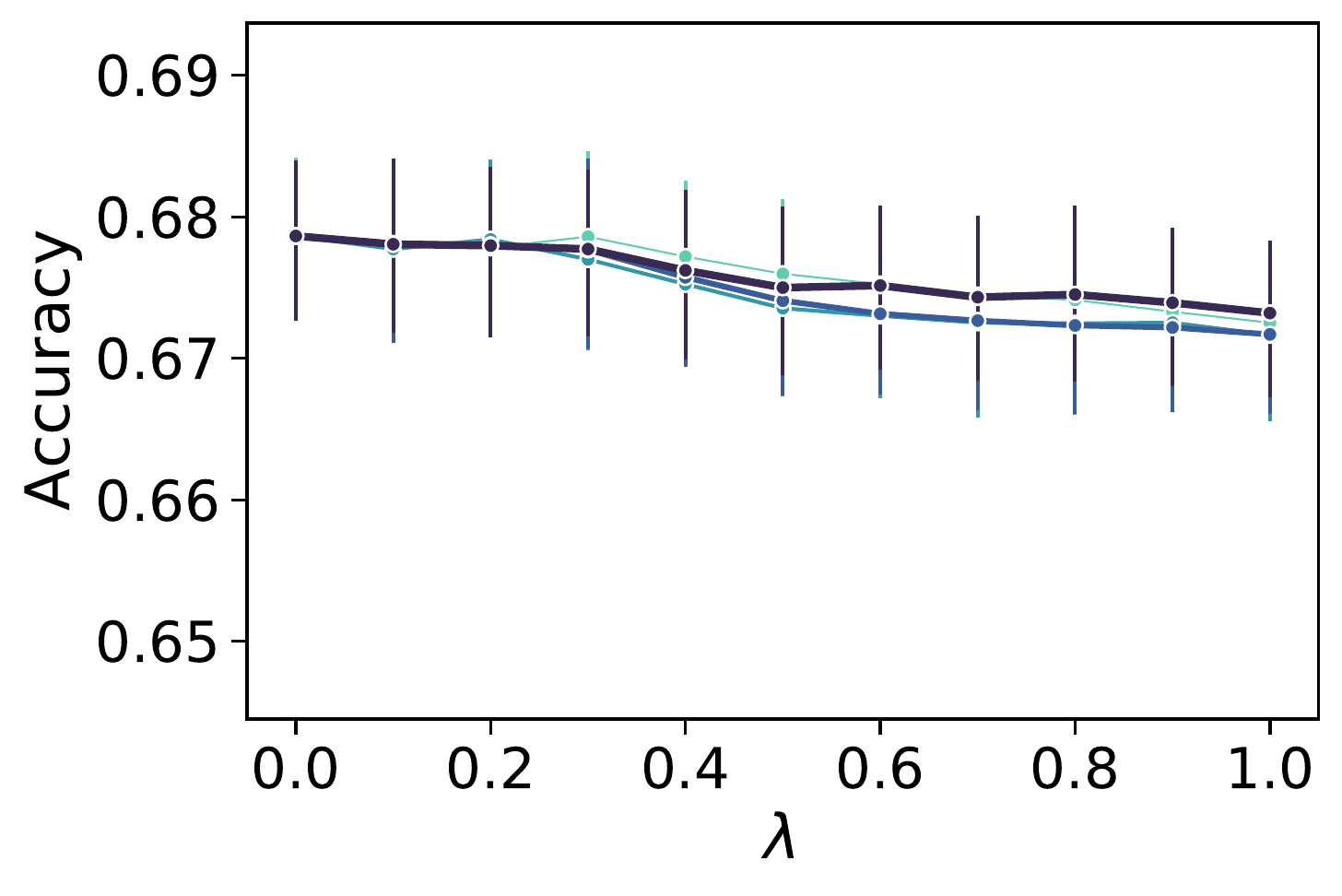}}
\\
\hspace{0.9cm}
{\includegraphics[width=0.4\linewidth]{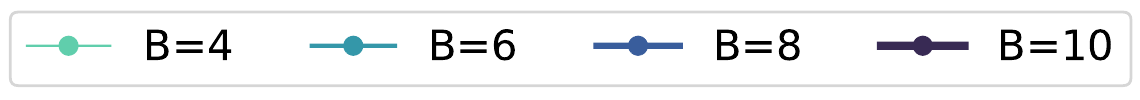}}
\caption{The effect of the repair tuning parameter $\lambda$ on distance, unfairness, and accuracy.
}
\label{fig:res_lam}
\end{figure*}

\subsection{An SMC Algorithm for Enhancing Fairness}\label{S:33}

In Section \ref{S:preprocess_mechanism} we described our fairness-enhancing mechanism, when all of $D$ is held by a single party (the centralized setting).
We now revisit that algorithm, and devise a secure implementation of it, given that the dataset $D$ is horizontally distributed among $L$ parties, as described in Section \ref{S31}.

\journal{
To that end, let us go back to Figure \ref{fig:method_basic}.
}
The steps that pose a challenge when privacy is of concern are the first two: dividing the non-sensitive attribute into equal-sized bins for each group $S \in \{U,V\}$, i.e., 
$D^S(X) = \bigcupdot_{i=1}^B b_{(i)}^S  $,
and then computing the boundaries of those bins,
see Eqs. (\ref{midef})-(\ref{mB1def}).
Performing those computations poses a challenge in the distributed setting, since they depend on data that is distributed among the $L$ parties and cannot be shared due to privacy concerns. 

The third step,
\journal{
in Figure \ref{fig:method_basic} 
}
preforming the repair,
poses no problem since it can be carried out by each party locally, independently of others, once the computed bin boundaries $m^S_{(i)}$, $i \in [B+1]$, $S \in \{U,V\}$, become known to each of the parties.
Even though in that step the parties do not collaborate, they must agree upfront on $\lambda$ (the repair tuning parameter), see Eq. (\ref{eq:repproposedII}).
Note that at this step, every $P_\ell$, $\ell \in [L]$, repairs the values of $D^V_\ell(X)$ that it possesses, but it does not share the repaired values with anyone else.

As for the fourth and last step,
\journal{
in Figure \ref{fig:method_basic}, there
}
training an ML classifier,
the parties need
to learn a classifier on distributed data, without sharing that data.
Since this is, again, a computational problem that involves all of the distributed dataset, privacy issues kick in. However, for such problems of privacy-preserving distributed ML classification
there are existing SMC-based solutions,  \cite[e.g.][]{lindell2000privacy,slavkovic2007secure,samet2015privacy,kikuchi2016efficient,fienberg2006secure}. 

Therefore, we focus on the problem of privacy-preserving binning of a distributed dataset, $D^S(X)$. We assume that all $L$ parties, $P_1,\ldots,P_L$,
agreed upfront on the number of bins, $B$, and that they
know the overall size of the dataset in each group, $n^S$, $S \in \{U,V\}$.
As $n^S=\sum_{\ell \in [L]} n^S_\ell $ and $n^S_\ell$ is known to $P_\ell$, 
$ \ell \in [L]$, then $n^S$ may be computed by a 
\textit{secure summation} sub-protocol \cite{clifton2002tools,benaloh1986secret,shi2011privacy}.

The privacy-preserving version of our fairness-enhancing mechanism assumes performing computations over integer values.
To be able to handle real values, we apply the following simple procedure to convert real values into integers (with some precision loss), prior to executing our mechanism.
If we are interested in preserving a precision of $d$ digits after the decimal point, we multiply each real value by $10^d$ and round the resulting value to the nearest integer.
After executing our mechanism, we divide the values by $10^d$.

Note that hereinafter, whenever we speak of $D(X)$ (or $D_\ell(X)$, $D^S(X)$, $D_\ell^S(X)$), we consider the values in that sensitive attribute after they were converted to integers, as described above.
We also assume that all parties know a lower and upper bound on the values of $D(X)$, denoted $\alpha$ and $\beta$ respectively. Namely, all entries in $D(X)$ (after they were converted to integers), are within the interval $[\alpha,\beta]$.

\journal{
Let us begin by recalling the principles of our binning scheme, as we described it earlier in Section \ref{S:preprocess_mechanism}.
Let 
$$ D^S(X)= (x_1,\ldots,x_{n^S})\,, \quad \mbox{where}~~
    x_1 \leq x_2 \leq \cdots \leq x_{n^S}\,,$$
be an ordering of the multiset of values in $D$'s attribute $X$ over group $S$, $D^S(X)$ (see Eq. (\ref{DSXorder})), and
let $K^S$ be the sequence of increasing indices in the range $[0,n^S]$, as defined in Eqs. (\ref{KS})-(\ref{kidef}).
Then the bins are given by
$$
    b^S_{(i)}:=\{ x_{j}: k_{(i-1)}+1 \leq j \leq k_{(i)}\} \,, \quad i \in [B]\,, $$
see Eq. (\ref{bindef}).
}
As $n^S$ and $B$ are both known to all parties,
everyone can compute 
the sequence $K^S$ of increasing indices in the range $[0,n^S]$, as defined in Eqs. (\ref{KS})-(\ref{kidef}).
Therefore, it is needed to compute, in a privacy-preserving manner, the $(k+1)$-th ranked element in $D^S(X)$ for every $k= k_{(i-1)}$ and $i \in [B]$, $S \in \{U,V\}$.
An SMC protocol for the privacy-preserving solution of this computational problem was introduced by
 \citeauthor{aggarwal2010secure} \shortcite{aggarwal2010secure}.

The solution in \citeauthor{aggarwal2010secure} \shortcite{aggarwal2010secure} relies on standard cryptographic building blocks of \textit{secure comparison} \cite{yao1982protocols} and \textit{secure summation}.
Secure comparison can be applied using generic protocols for SMC, such as \cite{goldreich1987play,franklin1992communication,beaver1990round}, while secure summation is a very simple computation, see e.g.  \cite{clifton2002tools,benaloh1986secret}. 
Hence, we can implement the protocol in \cite{aggarwal2010secure}
on top of standard libraries \cite[such as][]{damgaard2009asynchronous}.

The protocol in \cite{aggarwal2010secure} takes an iterative approach.
Assume that the parties wish to find the value of the
$k$-th ranked element, for some publicly known $k$. They perform that search iteratively, applying a binary-search approach.
As $\alpha$ and $\beta$ are the known lower and upper bounds on all values in $D(X)$, the first ``guess" for the value of the
$k$-th ranked element is $g:=\lfloor (\alpha+\beta)/2 \rfloor$.
Each $P_\ell$, $\ell \in [L]$, counts how many elements in $D^S_\ell(X)$ are smaller than $g$.
Then, by applying a secure
summation sub-protocol,
combined with a secure comparison sub-protocol,
they find out whether the number of elements smaller than $g$ in the unified dataset $D^S(X)$ is smaller than $k$ or not.
Based on the result of this iteration,  
the range is trimmed and a new guess is computed in the next iteration.

\begin{figure*}[t]
\settoheight{\tempheight}{\includegraphics[width=0.32\linewidth]{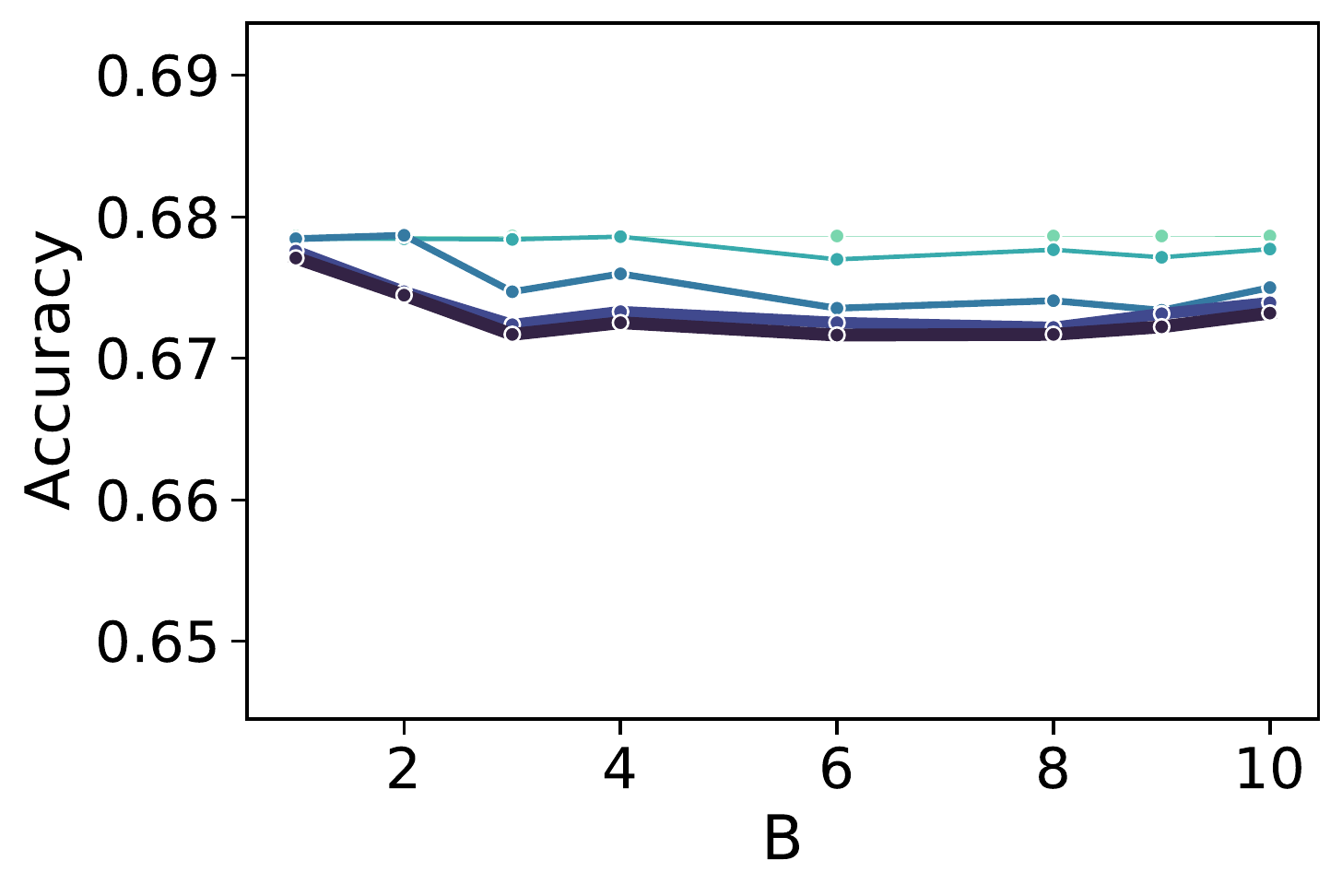}}
\centering
\hspace{\baselineskip}
\columnname{Distance}\hfil\hfil
\columnname{Unfairness}\hfil\hfil
\columnname{Accuracy}\hfil\hfil\\
\vspace{1mm}
{\includegraphics[width=0.305\linewidth]{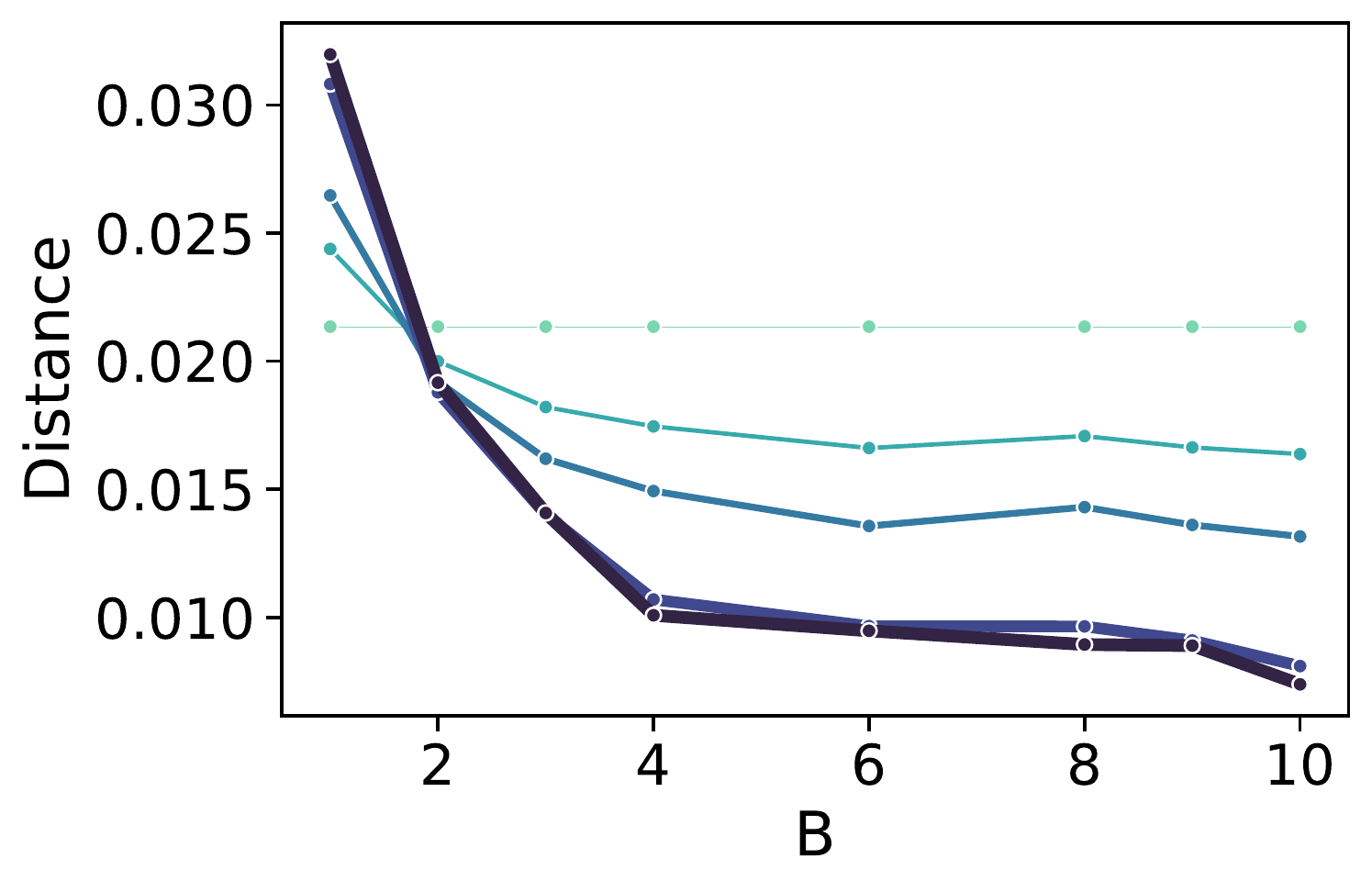}}
\hfil
{\includegraphics[width=0.29\linewidth]{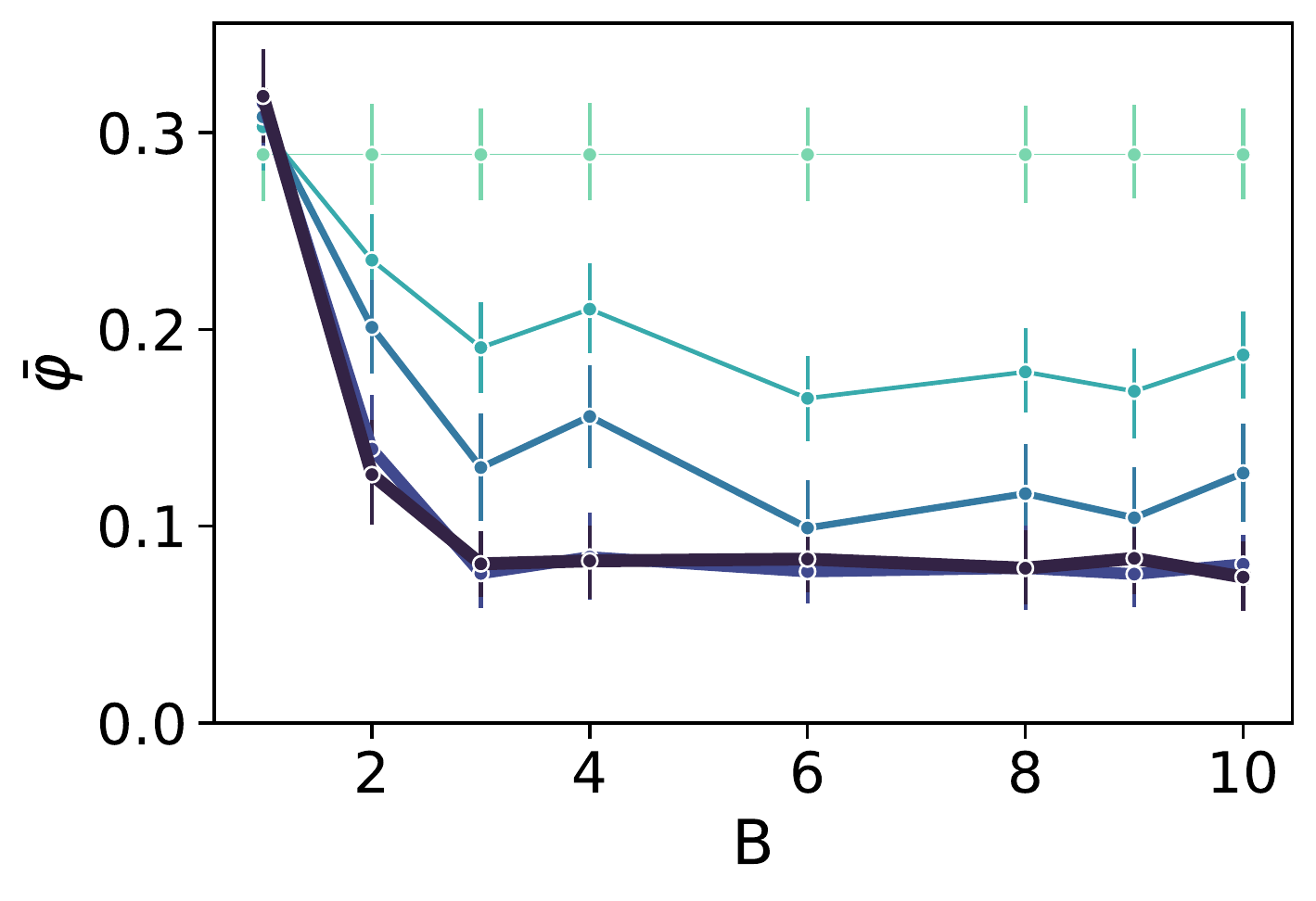}}
\hfil
{\includegraphics[width=0.3\linewidth]{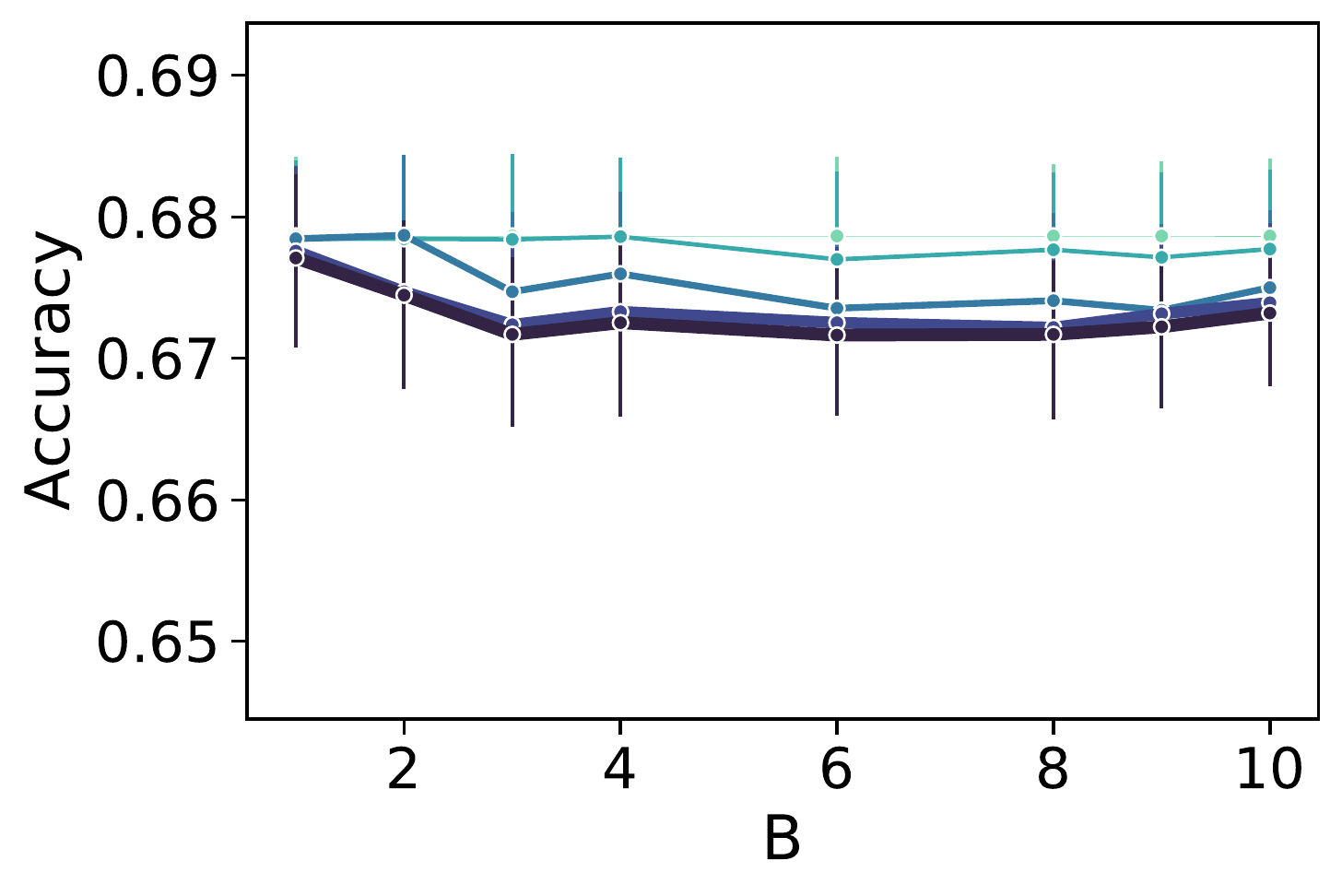}}
\\
\hspace{1cm}
{\includegraphics[width=0.48\linewidth]{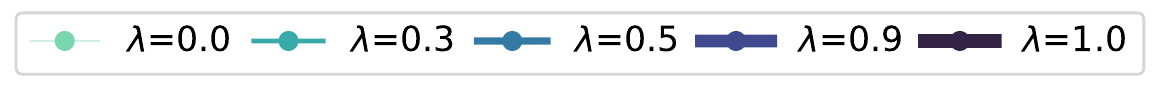}}
\caption{The effect of the number of bins $B$ on distance, unfairness, and accuracy.
}
\label{fig:res_bins}
\end{figure*}


\section{Evaluation}
\label{S:evaluation}



\journalC{In this section we present the experimental setting in terms of the dataset, the measures we used, and the parameters' space, and we also provide several implementation details.}

\journalC{\subsection*{Dataset}}
We evaluated the proposed method using the real-world publicly available ProPublica Recidivism dataset.
This dataset includes data from the COMPAS risk assessment system \cite[see][]{Angwin:2016:Online,Larson:2016:Online}.
The dataset includes 10 attributes, such as the number of previous felonies, charge degree, age, race, gender etc., and it has 6167 individual records.
The target attribute indicates whether an individual has recidivated (namely, was arrested again) after two years or not. The sensitive attribute in this dataset is race. All Caucasians constitute the privileged group, while all individuals of other races
constitute the unprivileged group.
\journalC{\subsection*{Measures}
As mentioned, there is an inherent trade-off between prediction performance and fairness.
We wish to evaluate how this trade-off is reflected in the application of our method.
To that end, we use the measures that are described below.
}
\journalC{\subsubsection*{Accuracy}}

Similarly to many studies on algorithmic fairness, we use accuracy as a measure of prediction performance. Accuracy is measured by the proportion of correct classifications.
\journalC{
\subsubsection*{Fairness}}
For measuring fairness, we consider a measure based on {equalized odds}.
As mentioned in Section \ref{S:algof}, one advantage of this measure
(in contrast, for example, to demographic parity), is that a perfectly accurate
classifier will be considered fair. 
More specifically, we define the unfairness measure 
$\bar{\varphi}=|DFNR|+|DFPR|$,
where $|DFNR|$ (resp. $|DFPR|$) is the absolute difference between the FNR (resp. FPR) of the two groups, and FNR (resp. FPR) is the False Negative Rate (resp. False Positive Rate).
\journalC{}
\journalB{
To that end, consider the confusion matrix shown in Table \ref{tab:confmat}:

\begin{table}[H]
\centering
\caption{Confusion matrix}
\label{tab:confmat}
\begin{tabular}{ccccl}
 &  & \multicolumn{2}{c}{\textbf{Predicted}} &  \\ \cline{3-4}
 & \multicolumn{1}{c||}{} & \multicolumn{1}{c|}{\textbf{"N"}} & \multicolumn{1}{c|}{\textbf{"P"}} &  
 \\\hhline{~===}
\multicolumn{1}{c|}{\multirow{2}{*}{\textbf
{Actual}}} & \multicolumn{1}{c||}{\textbf{N}} & \multicolumn{1}{c|}{TN} & \multicolumn{1}{c|}{FP} &  
\\ \cline{2-4}
\multicolumn{1}{c|}{} & \multicolumn{1}{c||}{\textbf{P}} & \multicolumn{1}{c|}{FN} & \multicolumn{1}{c|}{TP} &  \\ \cline{2-4}
\end{tabular}
\end{table}

Similarly to many studies on algorithmic fairness, we use accuracy as a measure of prediction performance. Accuracy is measured by the proportion of correct classifications, i.e. $(TN+TP)/(N+P)$.

\subsubsection*{Fairness}
For measuring fairness, we consider a measure based on {equalized odds}.
As mentioned in Section \ref{S:algof}, one advantage of this measure
(in contrast, for example, to demographic parity), is that a perfectly accurate
classifier will be considered fair. 
More specifically, we define the unfairness measure 
$\bar{\varphi}=|DFNR|+|DFPR|$,
where $|DFNR|$ (resp. $|DFPR|$) is the absolute difference between the FNR (resp. FPR) of the two groups, and FNR (resp. FPR) is the False Negative Rate (resp. False Positive Rate) as defined below:
$$ FNR=\frac{FN}{P}=\frac{FN}{FN+TP}\,,
\quad 
FPR=\frac{FP}{N}=\frac{FP}{FP+TN}\,.
$$ 
} 
\journalC{}
While the equalized odds measure dictates two separate measures, we use the sum of their absolute values in order to obtain a single combined measure.
Such a combined measure will allow us an easier examination of the fairness-accuracy trade-off at later stages, where higher values of $\bar{\varphi}$ indicate lower levels of fairness (or higher levels of unfairness).
\journalC{
}
\journalC{Note that in the ProPublica dataset, the value of "recidivated" is considered as the ``positive" value.
}
\journal{
\medskip
Note that in the two ProPublica datasets, the value of "recidivated" is considered as the ``positive" value, while in the case of the Bank Marketing dataset, a ``positive" value indicates that the corresponding individual has not subscribed to a term deposit service.
}
\journalC{
\subsubsection*{Distance}}
To better understand the repairing mechanism of our method, we also measure the distances between the distributions of attributes within the two groups. We do so by computing the \textit{earth mover’s distance} (EMD) \citep{rubner1998metric}, divided by $M$ (size of the range of possible values of the attribute).
\journal{
, see Section \ref{computational_costs}
}

For our experiments, we examined varying values of parameters, as follows:
\textit{Number of bins:} $B \in \{1,2,3,4,6,8,10\}$;
\textit{Repair tuning parameter:} $\lambda \in \{0.1,0.2,\ldots,0.9,1\}$; and
\textit{Number of parties:} in the majority of our experiments we applied a procedure based on three parties ($L=3$), whereas for the sake of measuring runtimes, we used higher numbers of parties ($L \in \{3,4,5,6,7\}$).
Our ML classifier was logistic regression.
In all experiments we used
a fixed number of $d=4$ digits after the decimal point,
and the same ratio in splitting the dataset into train set (66.7\%) and test set (33.3\%). Splits were repeated 10 different times in a random manner, and the reported results are the average and the 90\%-confidence interval over these 10 repetitions.
In each of the 10 repetitions we used the number $m$ of the repetition, $m \in \{0\ldots9\}$, as a random seed\footnote{Using {\it random.seed} function from {\it numpy} package.} for shuffling the dataset records and creating a new train-test split.

\journalC{\subsection*{Parameters' Space}

For our experiments, we examined varying values of parameters, as follows:
\begin{itemize}
    \item \textit{Number of bins:} $B \in \{1,2,3,4,6,8,10\}$.
    \item \textit{Repair tuning parameter:} $\lambda \in \{0.1,0.2,\ldots,0.9,1\}$.
    \item \textit{Number of parties:} in the majority of our experiments we applied a procedure based on three parties ($L=3$); only for the sake of measuring runtimes, we used higher numbers of parties ($L \in \{3,4,5,6,7\}$).
\end{itemize}

\smallskip\noindent
In all of our experiments we used:
\begin{itemize}
    \item A fixed number of $d=4$ digits after the decimal point.
    \item Logistic Regression as the ML classifier.
    \item A constant ratio to split the dataset into train set (66.7\%) and test set (33.3\%). Splits were repeated 10 different times in a random manner, and the reported results are the average over these 10 repetitions.
\end{itemize}
}

Our method was implemented in Python\footnote{The source code is attached in a zipped folder
as a supplementary material to this submission.}, assisted with the 
VIFF
library for secure multi-party computations \citep[see][]{damgaard2009asynchronous}.
For our purpose herein -- to examine the effects of our pre-process method, we used a simple {\it non-distributed non-private} implementation of the ML algorithm\footnote{We used {\it LogisticRegression} classifier from {\it sklearn} package with the parameters: $penalty="\ell_2"$ and $max\_iter=1000$.}.
All experiments were executed on a server running Windows Server 2008 R2, having two 6-cores CPU processors
with a clock speed of 1.9GHz, and 128GB of RAM.

\journalC{\subsection*{Implementation Details}
Our method was implemented in Python assisted with the Virtual Ideal Functionality Framework (VIFF) library for secure multi-party computations \citep[see][]{damgaard2009asynchronous}.
As noted above, our method was applied separately on each of the non-sensitive attributes of the considered dataset, prior to the training of an ML model.
For our purpose herein -- to examine the effects of our pre-process method, we used a simple {\it non-distributed non-private} implementation of the ML algorithm.}

\journalB{
All experiments were executed on a server running Windows Server 2008 R2, having two 6-cores CPU processors (12 virtual CPUs each) with a clock speed of 1.9GHz, and 128GB of RAM.
} 

\label{S:results}

\medskip
{\bf Results.}
We first evaluated the effect of our method on unfairness and accuracy.
In order to do so, we executed the method using varying
values of $\lambda$ and $B$, as mentioned in the previous section, and measured the resulting unfairness and accuracy values.
To better understand the repairing mechanism of our method, we also measured the resulting distance between the distributions of attributes within the two groups.
Recall that accuracy was measured by the proportion of correct classifications, unfairness by $\bar{\varphi}$, and distance by EMD.

Figure \ref{fig:res_lam} shows the effect of the repair tuning parameter $\lambda$ on the three considered measures for the real-world dataset.
Each chart represents a different measure: distance (left), unfairness (center) and accuracy (right).
In each chart, the $x$-axis represents $\lambda$, while the $y$-axis represents the value of the corresponding measure.
Colors and line thickness represent the value of $B$, where thicker lines represent higher values of $B$. 
Note that the distances are calculated for each attribute separately and are then averaged over the set of attributes in each dataset.
For all three measures, the reported results represent an average over 10 train-test splits. 
The vertical bars represent a 90\%-confidence interval.
As can be seen from the figure, by using the proposed method with higher values of $\lambda$, it is possible to improve fairness considerably with only a minor compromise in accuracy.
The considerable reduction in unfairness is a result of reducing the distances between the distributions of attributes within the two groups, as shown in the left chart in Figure \ref{fig:res_lam}.
For example, unfairness is reduced from 0.29 ($\lambda=0.0$) to 0.08 ($\lambda=1.0$) --- a reduction by 72\%; this is achieved with almost no compromise in accuracy (a decrease of less than 1\%).

Figure \ref{fig:res_bins} presents a similar analysis to the one presented in Figure \ref{fig:res_lam} to assess the effect of the number of bins $B$ on the three considered measures.
Here, the $x$-axis of each chart represents the value of $B$, while the $y$-axis represents the value of the corresponding measure.
Colors and line thickness represent the value of $\lambda$,
where thicker lines represent higher values of $\lambda$.
As can be seen from the figure, when increasing $B$, unfairness is reduced with only a minor compromise in accuracy.
However, increasing $B$ beyond $B=3$ has almost no effect on all three measures.
In particular, it barely contributes to the decrease in unfairness.
For example, using 10 bins with $\lambda \in \{0.9,1\}$ obtains about the same results as with only 3 bins.
The latter analysis indicates that it is preferable to use a small number of bins, around $B=3$.
Larger number of bins does not contribute towards enhancing fairness, but it does entail higher computational and communication costs, as well as increased leakage of information
\journal{
, as discussed in Sections \ref{S:34} and \ref{computational_costs}.
}
.

\begin{figure}[t]
\settoheight{\tempheight}{\includegraphics[width=1\linewidth]{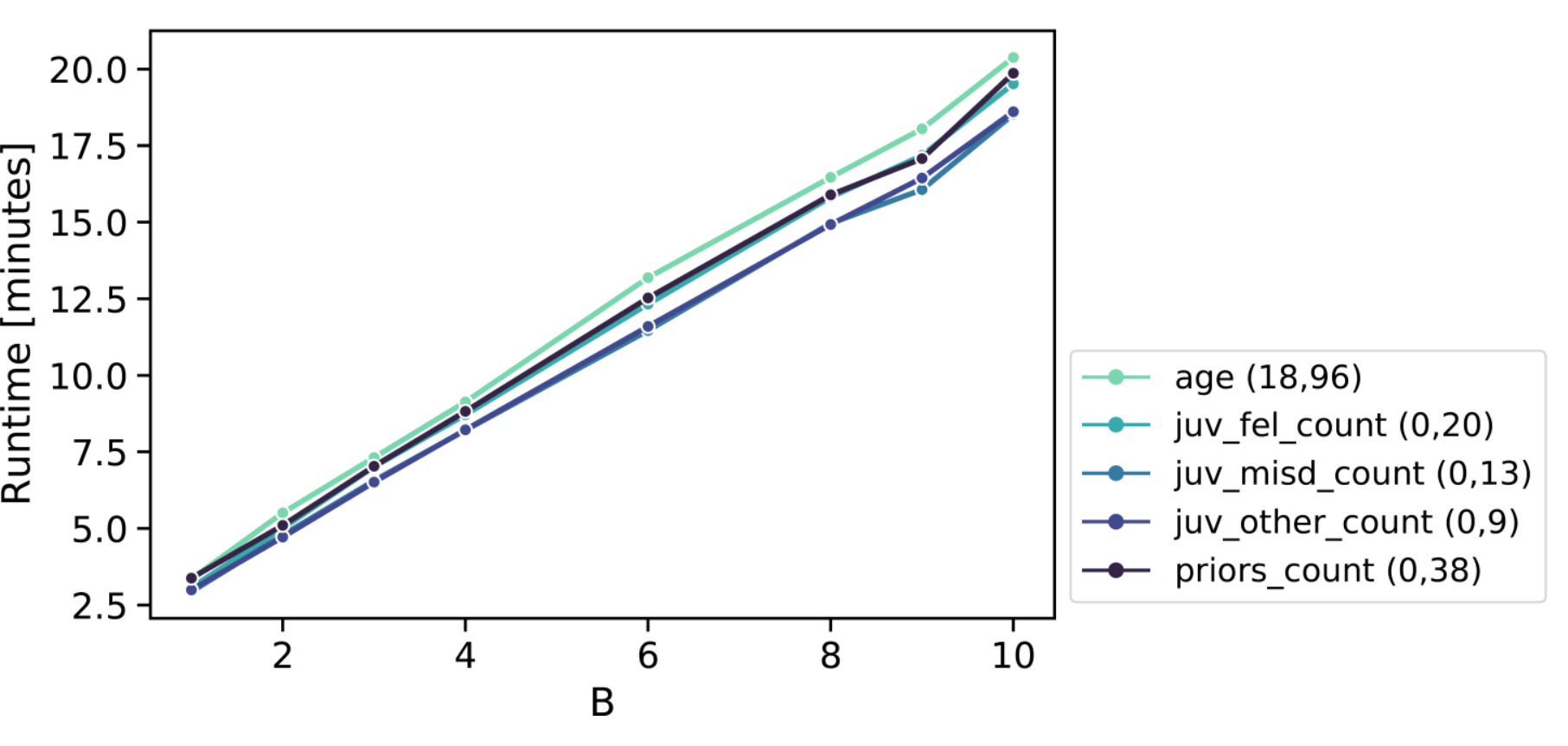}}%
\centering
    \includegraphics[width=0.72\linewidth]{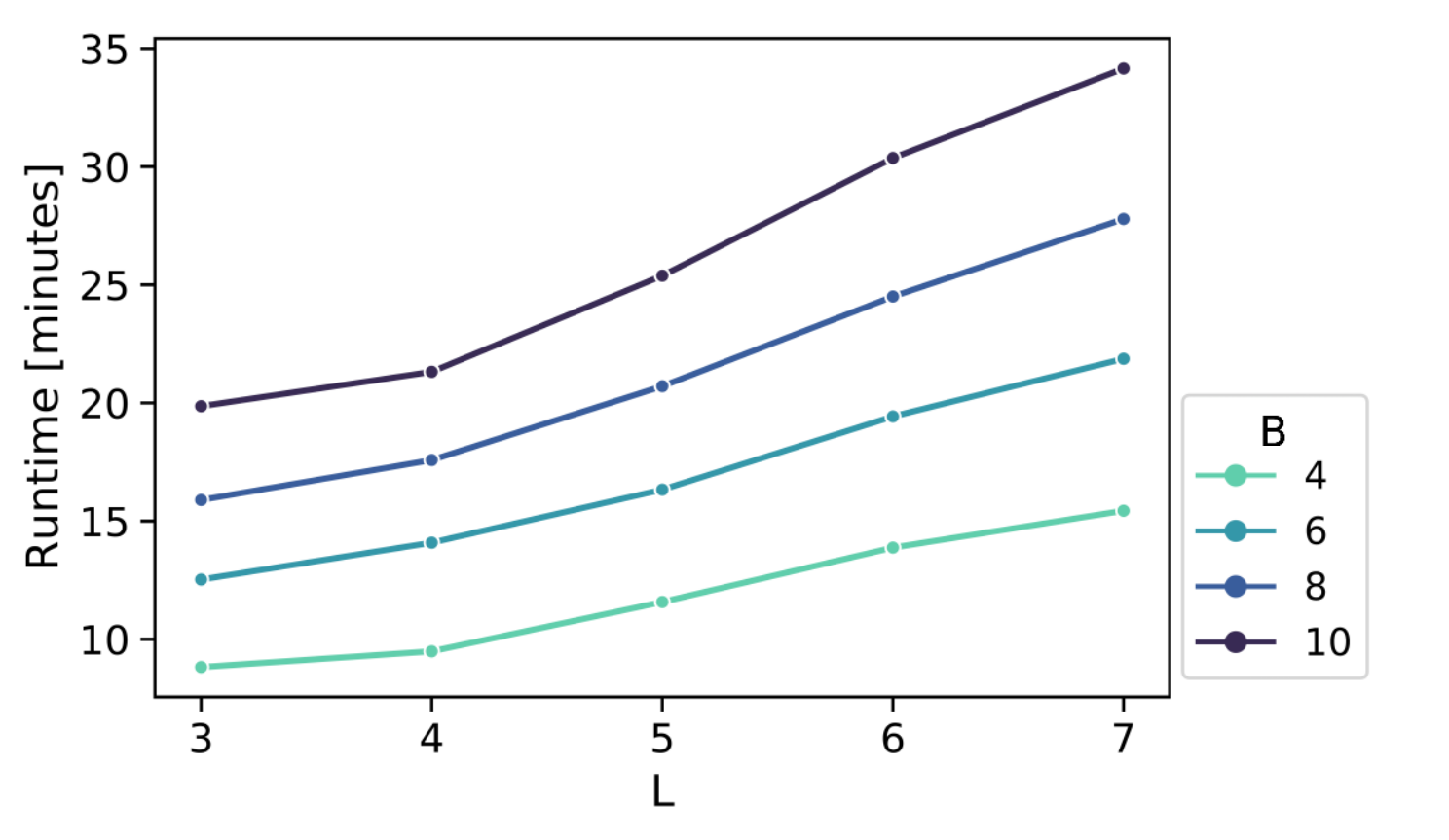}
\caption{The effect of parameters on runtime.}
\label{fig:runtimes}
\end{figure}

We then turned to evaluate the efficiency of our method, as reflected by its runtimes.
Figure \ref{fig:runtimes} shows the effect of $B$ (the number of bins) and $L$ (the number of parties) on the runtime of our method. Specifically, we report the runtime for the secure distributed computation of bin boundaries, for both of the groups.
The $x$-axis represents $L$, while the $y$-axis represents runtime in minutes.
Line colors represent the number of bins.
For clarity, we present the runtime for repairing one non-sensitive attribute - the ``prior count'' attribute.
The figure shows that
the runtime depends linearly on $L$ and linearly on $B$.
It is essential to note that the runtimes of the proposed method 
\journalC{(as demonstrated in Figures \ref{fig:runtime_bins} 
and \ref{fig:runtime_parties})}
(as demonstrated in Figure \ref{fig:runtimes})
are practical, especially considering that it is applied once.

To conclude, in the above experiments we showed that our method is able to improve fairness considerably with only a minor compromise in accuracy, despite their inherent trade-off.
We further showed that privacy is highly maintained and that the information leakage is minimal, especially as it appears that low values of $B$ suffice for fairness sake.
Finally, we showed that the runtime of the proposed method is feasible for a one-time pre-process procedure.

\section{Conclusions}
\label{S:conclusions}
We proposed herein a privacy-preserving pre-process mechanism for enhancing fairness of collaborative ML algorithms.
Our method improves fairness by decreasing distances between the distributions of attributes of the privileged and unprivileged groups. We use a binning approach that enables the implementation of privacy-preserving enhancements by means of SMC.
As a pre-process mechanism, our method is not limited to a specific algorithm, and hence it can be used with any collaborative ML algorithm.

The evaluation that we conducted, using a real-world dataset, revealed that the proposed method is able to improve fairness considerably, with only a minor compromise in accuracy.
\journal{
An extensive evaluation that was conducted using a real-world dataset, revealed that the proposed method is able to improve fairness considerably, with only a minor compromise in accuracy.
}
We also observed that using a small number of bins (e.g., $B=3$), it is possible to achieve that considerable improvement in fairness, with very minor and benign leakage of information.
\journalB{
We also showed that using a small number of bins (e.g., $B=3$), it is possible to achieve that considerable improvement in fairness, 
}
Finally, we demonstrated that the runtime of the proposed method is practical, especially considering that it is executed once as a pre-process procedure.

As future research, we suggest performing a broader analysis on the effect of the different parameters on the mechanism's computational costs. 
We also suggest to generalize the techniques presented here to handle non-binary sensitive attributes, such as age, with different age groups, race, or residential area. In addition, the current technique 
handles only numerical non-sensitive attributes; extending the technique to cope also with categorical attributes is in order. Finally, while we focused here on a horizontal distribution, a comprehensive discussion should include
a more general distribution framework.
Collaborative machine learning is encountered in many application scenarios, and questions of privacy and fairness naturally arise in all of them; hence, we see the current study as a first step in a long quest.



\begin{thebibliography}{57}
\providecommand{\natexlab}[1]{#1}

\bibitem[{Abadi et~al.(2016)Abadi, Chu, Goodfellow, McMahan, Mironov, Talwar,
  and Zhang}]{abadi2016deep}
Abadi, M.; Chu, A.; Goodfellow, I.; McMahan, H.~B.; Mironov, I.; Talwar, K.;
  and Zhang, L. 2016.
\newblock Deep learning with differential privacy.
\newblock In \emph{Proceedings of the 2016 ACM SIGSAC Conference on Computer
  and Communications Security}, 308--318.

\bibitem[{Agarwal et~al.(2018)Agarwal, Beygelzimer, Dudik, Langford, and
  Wallach}]{agarwal2018reductions}
Agarwal, A.; Beygelzimer, A.; Dudik, M.; Langford, J.; and Wallach, H. 2018.
\newblock A reductions approach to fair classification.
\newblock In \emph{International Conference on Machine Learning}, 60--69.

\bibitem[{Aggarwal, Mishra, and Pinkas(2010)}]{aggarwal2010secure}
Aggarwal, G.; Mishra, N.; and Pinkas, B. 2010.
\newblock Secure computation of the median (and other elements of specified
  ranks).
\newblock \emph{Journal of cryptology}, 23: 373--401.

\bibitem[{Agrawal and Srikant(2000)}]{agrawal2000privacy}
Agrawal, R.; and Srikant, R. 2000.
\newblock Privacy-preserving data mining.
\newblock In \emph{Proceedings of the 2000 ACM SIGMOD international conference
  on Management of data}, 439--450.

\bibitem[{Angwin(2016)}]{Angwin:2016:Online}
Angwin, J. 2016.
\newblock Machine bias — ProPublica.

\bibitem[{Bagdasaryan, Poursaeed, and
  Shmatikov(2019)}]{bagdasaryan2019differential}
Bagdasaryan, E.; Poursaeed, O.; and Shmatikov, V. 2019.
\newblock Differential privacy has disparate impact on model accuracy.
\newblock In \emph{Advances in Neural Information Processing Systems},
  15479--15488.

\bibitem[{Beaver, Micali, and Rogaway(1990)}]{beaver1990round}
Beaver, D.; Micali, S.; and Rogaway, P. 1990.
\newblock The round complexity of secure protocols.
\newblock In \emph{Proceedings of the twenty-second annual ACM symposium on
  Theory of computing}, 503--513.

\bibitem[{Benaloh(1986)}]{benaloh1986secret}
Benaloh, J.~C. 1986.
\newblock Secret sharing homomorphisms: Keeping shares of a secret secret.
\newblock In \emph{Conference on the Theory and Application of Cryptographic
  Techniques}, 251--260. Springer.

\bibitem[{Calders and Verwer(2010)}]{calders2010three}
Calders, T.; and Verwer, S. 2010.
\newblock Three naive Bayes approaches for discrimination-free classification.
\newblock \emph{Data Mining and Knowledge Discovery}, 21: 277--292.

\bibitem[{Calmon et~al.(2017)Calmon, Wei, Vinzamuri, Ramamurthy, and
  Varshney}]{calmon2017optimized}
Calmon, F.; Wei, D.; Vinzamuri, B.; Ramamurthy, K.~N.; and Varshney, K.~R.
  2017.
\newblock Optimized pre-processing for discrimination prevention.
\newblock In \emph{Advances in Neural Information Processing Systems},
  3992--4001.

\bibitem[{Chase et~al.(2017)Chase, Gilad-Bachrach, Laine, Lauter, and
  Rindal}]{chase2017private}
Chase, M.; Gilad-Bachrach, R.; Laine, K.; Lauter, K.~E.; and Rindal, P. 2017.
\newblock Private collaborative neural network learning.
\newblock \emph{IACR Cryptol. ePrint Arch.}, 2017: 762.

\bibitem[{Clifton et~al.(2002)Clifton, Kantarcioglu, Vaidya, Lin, and
  Zhu}]{clifton2002tools}
Clifton, C.; Kantarcioglu, M.; Vaidya, J.; Lin, X.; and Zhu, M.~Y. 2002.
\newblock Tools for privacy preserving distributed data mining.
\newblock \emph{ACM Sigkdd Explorations Newsletter}, 4: 28--34.

\bibitem[{Corbett-Davies et~al.(2017)Corbett-Davies, Pierson, Feller, Goel, and
  Huq}]{corbett2017algorithmic}
Corbett-Davies, S.; Pierson, E.; Feller, A.; Goel, S.; and Huq, A. 2017.
\newblock Algorithmic decision making and the cost of fairness.
\newblock In \emph{Proceedings of the 23rd ACM SIGKDD International Conference
  on Knowledge Discovery and Data Mining}, 797--806. ACM.

\bibitem[{Cummings et~al.(2019)Cummings, Gupta, Kimpara, and
  Morgenstern}]{cummings2019compatibility}
Cummings, R.; Gupta, V.; Kimpara, D.; and Morgenstern, J. 2019.
\newblock On the compatibility of privacy and fairness.
\newblock In \emph{Adjunct Publication of the 27th Conference on User Modeling,
  Adaptation and Personalization}, 309--315.

\bibitem[{Damg{\aa}rd et~al.(2009)Damg{\aa}rd, Geisler, Kr{\o}igaard, and
  Nielsen}]{damgaard2009asynchronous}
Damg{\aa}rd, I.; Geisler, M.; Kr{\o}igaard, M.; and Nielsen, J.~B. 2009.
\newblock Asynchronous multiparty computation: Theory and implementation.
\newblock In \emph{International workshop on public key cryptography},
  160--179. Springer.

\bibitem[{Domingo-Ferrer and Blanco-Justicia(2020)}]{domingo2020privacy}
Domingo-Ferrer, J.; and Blanco-Justicia, A. 2020.
\newblock Privacy-preserving technologies.
\newblock In \emph{The Ethics of Cybersecurity}, 279--297. Springer, Cham.

\bibitem[{Du, Han, and Chen(2004)}]{du2004privacy}
Du, W.; Han, Y.~S.; and Chen, S. 2004.
\newblock Privacy-preserving multivariate statistical analysis: Linear
  regression and classification.
\newblock In \emph{Proceedings of the 2004 SIAM international conference on
  data mining}, 222--233. SIAM.

\bibitem[{Dwork et~al.(2012)Dwork, Hardt, Pitassi, Reingold, and
  Zemel}]{dwork2012fairness}
Dwork, C.; Hardt, M.; Pitassi, T.; Reingold, O.; and Zemel, R. 2012.
\newblock Fairness through awareness.
\newblock In \emph{Proceedings of the 3rd innovations in theoretical computer
  science conference}, 214--226. ACM.

\bibitem[{Dwork et~al.(2018)Dwork, Immorlica, Kalai, and
  Leiserson}]{dwork2018decoupled}
Dwork, C.; Immorlica, N.; Kalai, A.~T.; and Leiserson, M. 2018.
\newblock Decoupled classifiers for group-fair and efficient machine learning.
\newblock In \emph{Conference on Fairness, Accountability and Transparency},
  119--133.

\bibitem[{Fantin(2020)}]{fantin2020distributed}
Fantin, J. 2020.
\newblock A distributed fair random forest.

\bibitem[{Feldman et~al.(2015)Feldman, Friedler, Moeller, Scheidegger, and
  Venkatasubramanian}]{feldman2015certifying}
Feldman, M.; Friedler, S.~A.; Moeller, J.; Scheidegger, C.; and
  Venkatasubramanian, S. 2015.
\newblock Certifying and removing disparate impact.
\newblock In \emph{Proceedings of the 21th ACM SIGKDD International Conference
  on Knowledge Discovery and Data Mining}, 259--268. ACM.

\bibitem[{Fienberg et~al.(2006)Fienberg, Fulp, Slavkovic, and
  Wrobel}]{fienberg2006secure}
Fienberg, S.~E.; Fulp, W.~J.; Slavkovic, A.~B.; and Wrobel, T.~A. 2006.
\newblock “Secure” log-linear and logistic regression analysis of
  distributed databases.
\newblock In \emph{International Conference on Privacy in Statistical
  Databases}, 277--290. Springer.

\bibitem[{Franklin and Yung(1992)}]{franklin1992communication}
Franklin, M.; and Yung, M. 1992.
\newblock Communication complexity of secure computation.
\newblock In \emph{Proceedings of the twenty-fourth annual ACM symposium on
  Theory of computing}, 699--710.

\bibitem[{Goldreich, Micali, and Wigderson(1987)}]{goldreich1987play}
Goldreich, O.; Micali, S.; and Wigderson, A. 1987.
\newblock How to play any mental game.
\newblock In \emph{Proceedings of the nineteenth annual ACM symposium on Theory
  of computing}, 218--229.

\bibitem[{Hardt, Price, and Srebro(2016)}]{hardt2016equality}
Hardt, M.; Price, E.; and Srebro, N. 2016.
\newblock Equality of opportunity in supervised learning.
\newblock In \emph{Advances in neural information processing systems},
  3315--3323.

\bibitem[{Hu et~al.(2019{\natexlab{a}})Hu, Liu, Wang, and
  Lan}]{hu2019distributed}
Hu, H.; Liu, Y.; Wang, Z.; and Lan, C. 2019{\natexlab{a}}.
\newblock A distributed fair machine learning framework with private
  demographic data protection.
\newblock In \emph{2019 IEEE International Conference on Data Mining (ICDM)},
  1102--1107. IEEE.

\bibitem[{Hu et~al.(2019{\natexlab{b}})Hu, Niu, Yang, and Zhou}]{hu2019fdml}
Hu, Y.; Niu, D.; Yang, J.; and Zhou, S. 2019{\natexlab{b}}.
\newblock FDML: A collaborative machine learning framework for distributed
  features.
\newblock In \emph{Proceedings of the 25th ACM SIGKDD International Conference
  on Knowledge Discovery \& Data Mining}, 2232--2240.

\bibitem[{Huang et~al.(2018)Huang, Chen, Kairouz, Sankar, and
  Rajagopal}]{huang2018generative}
Huang, C.; Chen, X.; Kairouz, P.; Sankar, L.; and Rajagopal, R. 2018.
\newblock Generative adversarial models for learning private and fair
  representations.

\bibitem[{Jagielski et~al.(2019)Jagielski, Kearns, Mao, Oprea, Roth,
  Sharifi-Malvajerdi, and Ullman}]{jagielski2019differentially}
Jagielski, M.; Kearns, M.; Mao, J.; Oprea, A.; Roth, A.; Sharifi-Malvajerdi,
  S.; and Ullman, J. 2019.
\newblock Differentially private fair learning.
\newblock In \emph{International Conference on Machine Learning}, 3000--3008.

\bibitem[{Jayaraman et~al.(2018)Jayaraman, Wang, Evans, and
  Gu}]{jayaraman2018distributed}
Jayaraman, B.; Wang, L.; Evans, D.; and Gu, Q. 2018.
\newblock Distributed learning without distress: Privacy-preserving empirical
  risk minimization.
\newblock In \emph{Advances in Neural Information Processing Systems},
  6343--6354.

\bibitem[{Jeckmans, Tang, and Hartel(2012)}]{jeckmans2012privacy}
Jeckmans, A.; Tang, Q.; and Hartel, P. 2012.
\newblock Privacy-preserving collaborative filtering based on horizontally
  partitioned dataset.
\newblock In \emph{2012 International Conference on Collaboration Technologies
  and Systems (CTS)}, 439--446. IEEE.

\bibitem[{Jha, Kruger, and McDaniel(2005)}]{jha2005privacy}
Jha, S.; Kruger, L.; and McDaniel, P. 2005.
\newblock Privacy preserving clustering.
\newblock In \emph{European symposium on research in computer security},
  397--417. Springer.

\bibitem[{Kamishima et~al.(2012)Kamishima, Akaho, Asoh, and
  Sakuma}]{kamishima2012fairness}
Kamishima, T.; Akaho, S.; Asoh, H.; and Sakuma, J. 2012.
\newblock Fairness-aware classifier with prejudice remover regularizer.
\newblock In \emph{Joint European Conference on Machine Learning and Knowledge
  Discovery in Databases}, 35--50. Springer.

\bibitem[{Kikuchi et~al.(2016)Kikuchi, Yasunaga, Matsui, and
  Fan}]{kikuchi2016efficient}
Kikuchi, H.; Yasunaga, H.; Matsui, H.; and Fan, C.-I. 2016.
\newblock Efficient privacy-preserving logistic regression with iteratively
  re-weighted least squares.
\newblock In \emph{2016 11th Asia Joint Conference on Information Security
  (AsiaJCIS)}, 48--54. IEEE.

\bibitem[{Kilbertus et~al.(2018)Kilbertus, Gasc{\'o}n, Kusner, Veale, Gummadi,
  and Weller}]{kilbertus2018blind}
Kilbertus, N.; Gasc{\'o}n, A.; Kusner, M.~J.; Veale, M.; Gummadi, K.~P.; and
  Weller, A. 2018.
\newblock Blind Justice: Fairness with Encrypted Sensitive Attributes.
\newblock In \emph{Proceedings of the 35th International Conference on Machine
  Learning (ICML 2018)}, volume~80, 2635--2644.

\bibitem[{Kleinberg, Mullainathan, and Raghavan(2017)}]{kleinberg2017inherent}
Kleinberg, J.; Mullainathan, S.; and Raghavan, M. 2017.
\newblock Inherent trade-Offs in the fair determination of risk scores.
\newblock In \emph{8th Innovations in Theoretical Computer Science Conference
  (ITCS 2017)}. Schloss Dagstuhl-Leibniz-Zentrum fuer Informatik.

\bibitem[{Krizhevsky, Sutskever, and Hinton(2012)}]{krizhevsky2012imagenet}
Krizhevsky, A.; Sutskever, I.; and Hinton, G.~E. 2012.
\newblock Imagenet classification with deep convolutional neural networks.
\newblock In \emph{Advances in neural information processing systems},
  1097--1105.

\bibitem[{Larson et~al.(2016)Larson, Mattu, Kirchner, and
  Angwin}]{Larson:2016:Online}
Larson, J.; Mattu, S.; Kirchner, L.; and Angwin, J. 2016.
\newblock How we analyzed the COMPAS recidivism algorithm.

\bibitem[{Lindell and Pinkas(2000)}]{lindell2000privacy}
Lindell, Y.; and Pinkas, B. 2000.
\newblock Privacy preserving data mining.
\newblock In \emph{Annual International Cryptology Conference}, 36--54.
  Springer.

\bibitem[{Louizos et~al.(2017)Louizos, Swersky, Li, Welling, and
  Zemel}]{louizos2017variational}
Louizos, C.; Swersky, K.; Li, Y.; Welling, M.; and Zemel, R. 2017.
\newblock The variational fair autoencoder.
\newblock \emph{arXiv preprint arXiv:1511.00830}.

\bibitem[{Mangasarian, Wild, and Fung(2008)}]{mangasarian2008privacy}
Mangasarian, O.~L.; Wild, E.~W.; and Fung, G.~M. 2008.
\newblock Privacy-preserving classification of vertically partitioned data via
  random kernels.
\newblock \emph{ACM Transactions on Knowledge Discovery from Data (TKDD)}, 2:
  1--16.

\bibitem[{Menon and Williamson(2018)}]{menon2018cost}
Menon, A.~K.; and Williamson, R.~C. 2018.
\newblock The cost of fairness in binary classification.
\newblock In \emph{Conference on Fairness, Accountability and Transparency},
  107--118.

\bibitem[{Mozannar, Ohannessian, and Srebro(2020)}]{mozannar2020fair}
Mozannar, H.; Ohannessian, M.~I.; and Srebro, N. 2020.
\newblock Fair learning with private demographic data.
\newblock \emph{arXiv preprint arXiv:2002.11651}.

\bibitem[{Pessach and Shmueli(2020)}]{pessach2020algorithmic}
Pessach, D.; and Shmueli, E. 2020.
\newblock Algorithmic fairness.
\newblock \emph{arXiv preprint arXiv:2001.09784}.

\bibitem[{Rubner, Tomasi, and Guibas(1998)}]{rubner1998metric}
Rubner, Y.; Tomasi, C.; and Guibas, L.~J. 1998.
\newblock A metric for distributions with applications to image databases.
\newblock In \emph{Sixth International Conference on Computer Vision (IEEE Cat.
  No. 98CH36271)}, 59--66. IEEE.

\bibitem[{Samet(2015)}]{samet2015privacy}
Samet, S. 2015.
\newblock Privacy-preserving logistic regression.
\newblock \emph{Journal of Advances in Information Technology Vol}, 6.

\bibitem[{Shi et~al.(2011)Shi, Chan, Rieffel, Chow, and Song}]{shi2011privacy}
Shi, E.; Chan, T.~H.; Rieffel, E.; Chow, R.; and Song, D. 2011.
\newblock Privacy-preserving aggregation of time-series data.
\newblock In \emph{Annual Network and Distributed System Security Symposium
  (NDSS)}, volume~2, 1--17. Internet Society.

\bibitem[{Slavkovic, Nardi, and Tibbits(2007)}]{slavkovic2007secure}
Slavkovic, A.~B.; Nardi, Y.; and Tibbits, M.~M. 2007.
\newblock "Secure" logistic regression of horizontally and vertically
  partitioned distributed databases.
\newblock In \emph{Seventh IEEE International Conference on Data Mining
  Workshops (ICDMW 2007)}, 723--728. IEEE.

\bibitem[{Song, Chaudhuri, and Sarwate(2013)}]{song2013stochastic}
Song, S.; Chaudhuri, K.; and Sarwate, A.~D. 2013.
\newblock Stochastic gradient descent with differentially private updates.
\newblock In \emph{2013 IEEE Global Conference on Signal and Information
  Processing}, 245--248. IEEE.

\bibitem[{Tassa, Grinshpoun, and Yanai(2021)}]{TassaGY21}
Tassa, T.; Grinshpoun, T.; and Yanai, A. 2021.
\newblock PC-SyncBB: {A} privacy preserving collusion secure {DCOP} algorithm.
\newblock \emph{Artif. Intell.}, 297: 103501.

\bibitem[{{U.S. Department of Health and Human
  Services}(1996)}]{HIPAA:1996:Online}
{U.S. Department of Health and Human Services}. 1996.
\newblock Summary of the HIPAA Security Rule.

\bibitem[{Verma and Rubin(2018)}]{verma2018fairness}
Verma, S.; and Rubin, J. 2018.
\newblock Fairness definitions explained.
\newblock In \emph{2018 IEEE/ACM International Workshop on Software Fairness
  (FairWare)}, 1--7. IEEE.

\bibitem[{Xu, Yuan, and Wu(2019)}]{xu2019achieving}
Xu, D.; Yuan, S.; and Wu, X. 2019.
\newblock Achieving differential privacy and fairness in logistic regression.
\newblock In \emph{Companion Proceedings of The 2019 World Wide Web
  Conference}, 594--599.

\bibitem[{Yang et~al.(2019)Yang, Liu, Chen, and Tong}]{yang2019federated}
Yang, Q.; Liu, Y.; Chen, T.; and Tong, Y. 2019.
\newblock Federated machine learning: Concept and applications.
\newblock \emph{ACM Transactions on Intelligent Systems and Technology (TIST)},
  10: 1--19.

\bibitem[{Yao(1982)}]{yao1982protocols}
Yao, A.~C. 1982.
\newblock Protocols for secure computations.
\newblock In \emph{23rd annual symposium on foundations of computer science
  (sfcs 1982)}, 160--164. IEEE.

\bibitem[{Zafar et~al.(2017)Zafar, Valera, Gomez~Rodriguez, and
  Gummadi}]{zafar2017fairness}
Zafar, M.~B.; Valera, I.; Gomez~Rodriguez, M.; and Gummadi, K.~P. 2017.
\newblock Fairness beyond disparate treatment \& disparate impact: Learning
  classification without disparate mistreatment.
\newblock In \emph{Proceedings of the 26th International Conference on World
  Wide Web}, 1171--1180. International World Wide Web Conferences Steering
  Committee.

\bibitem[{Zemel et~al.(2013)Zemel, Wu, Swersky, Pitassi, and
  Dwork}]{zemel2013learning}
Zemel, R.; Wu, Y.; Swersky, K.; Pitassi, T.; and Dwork, C. 2013.
\newblock Learning fair representations.
\newblock In \emph{International Conference on Machine Learning}, 325--333.

\end{thebibliography}
\fontsize{9.0pt}{10.0pt}\selectfont

\end{document}